\documentclass[twocolumn]{article}

\usepackage{PRIMEarxiv}

\usepackage[utf8]{inputenc} % allow utf-8 input
\usepackage[T1]{fontenc}    % use 8-bit T1 fonts
\usepackage{hyperref}       % hyperlinks
\usepackage{url}            % simple URL typesetting
\usepackage{booktabs}       % professional-quality tables
\usepackage{amsfonts}       % blackboard math symbols
\usepackage{nicefrac}       % compact symbols for 1/2, etc.
\usepackage{microtype}      % microtypography
\usepackage{lipsum}
\usepackage{subcaption}
\usepackage{amsmath}
\usepackage{dblfloatfix}
\usepackage{algorithm}
\usepackage{algpseudocode}
\usepackage{fancyhdr}       % header
\usepackage{graphicx}       % graphics
\graphicspath{{media/}}     % organize your images and other figures under media/ folder

\newcommand*{\img}[1]{%
    \raisebox{-.05\baselineskip}{%
        \includegraphics[
        height=.6\baselineskip,
        width=.6\baselineskip,
        keepaspectratio,
        ]{#1}%
    }%
}

\renewcommand{\thefootnote}{\fnsymbol{footnote}}
  
\title{Spatio-Temporal Wind Speed Forecasting using Graph Networks and Novel Transformer Architectures
}

\author{
  Lars Ødegaard Bentsen\thanks{See Below...} \\
  Department of Technology Systems \\
  University of Oslo \\ 
  Kjeller, Norway \\
  \texttt{lars.bentsen@uio.its.no} \\
  \And
  Narada Dilp Warakagoda \\
  Department of Technology Systems \\
  University of Oslo \\
  Kjeller, Norway \\
  \AND
  Roy Stenbro \\
  Institute for Energy Systems (IFE) \\
  Kjeller, Norway
  \And
  Paal Engelstad \\
  Department of Technology Systems \\
  University of Oslo \\
  Kjeller, Norway \\
}

\begin{document}

\twocolumn[%
%  Title and authors
  \maketitle
  \vspace{1ex}%
  \begin{abstract}
  \vspace{1ex}
    To improve the security and reliability of wind energy production, short-term forecasting has become of utmost importance.
  This study focuses on multi-step spatio-temporal wind speed forecasting for the Norwegian continental shelf. 
  In particular, the study considers 14 offshore measurement stations and aims to leverage spatial dependencies through the relative physical location of different stations to improve local wind forecasts and simultaneously output different forecasts for each of the 14 locations. 
  Our multi-step forecasting models produce either 10-minute, 1- or 4-hour forecasts, with 10-minute resolution, meaning that the models produce more informative time series for predicted future trends.
  A graph neural network (GNN) architecture was used to extract spatial dependencies, with different update functions to learn temporal correlations. 
  These update functions were implemented using different neural network architectures.
  One such architecture, the Transformer, has become increasingly popular for sequence modelling in recent years. 
  Various alterations have been proposed to better facilitate time series forecasting, of which this study focused on the Informer, LogSparse Transformer and Autoformer. 
  This is the first time the LogSparse Transformer and Autoformer have been applied to wind forecasting and the first time any of these or the Informer have been formulated in a spatio-temporal setting for wind forecasting.
  By comparing against spatio-temporal Long Short-Term Memory (LSTM) and Multi-Layer Perceptron (MLP) models, the study showed that the models using the altered Transformer architectures as update functions in GNNs were able to outperform these.
  Furthermore, we propose the Fast Fourier Transformer (FFTransformer), which is a novel Transformer architecture based on signal decomposition and consists of two separate streams that analyse the trend and periodic components separately.  
  The FFTransformer and Autoformer were found to achieve superior results for the 10-minute and 1-hour ahead forecasts, with the FFTransformer significantly outperforming all other models for the 4-hour ahead forecasts.
  Our code to implement the different models are made publicly available at: \url{https://github.com/LarsBentsen/FFTransformer}.
  \end{abstract}
  \vspace{2ex}
  \keywords{Spatio-Temporal Wind Forecasting \and Multi-Step \and Transformers \and Graph Neural Networks}
  \vspace{2ex}
] \footnotetext[1]{\textit{Corresponding Author}}

% \begin{abstract}
%     \lipsum[1]
% \end{abstract}
% \maketitle

% keywords can be removed
% \keywords{Spatio-Temporal \and Short-Term Forecasting \and Transformers \and Graph Neural Networks}

\renewcommand{\thefootnote}{\arabic{footnote}}
\setcounter{footnote}{0}

\section{Introduction}
In the context of the global climate debate, wind has emerged as a prominent renewable energy resource to accelerate the depletion of fossil fuel-based energy production \cite{okumus2016current}.
Nevertheless, in contrast to conventional power plants, wind resources are inherently intermittent in the short-term, which poses significant challenges for operators and grid planning \cite{van2016long}. 
To alleviate some of these and help facilitate large-scale adoption of wind power, accurate wind forecasting has become of critical importance.

Wind forecasting methods can be categorised as either physical or statistical. 
Physical models are based on detailed physical laws that model the atmosphere and typically aim to increase the resolution of coarse numerical weather prediction (NWP) models. 
A challenge with physical models is that they come at a very high computational cost, making them less viable for local short-term forecasting \cite{chang2014literature}.
Statistical and machine learning (ML) methods, on the other hand, leverage historical data to optimise model parameters.
Even though the training of ML models might be a time-exhaustive process, such models are very quick during inference, meaning that forecasts can be obtained in near real-time. 

This paper focuses on spatio-temporal multi-step wind forecasting based on recent developments in deep learning (DL). 
With multi-step forecasting, the study aims to output more informative time series. 
Considering a scenario of forecasting one hour ahead, some studies give a single prediction for the average wind speed over the entire period. 
However, sometimes one might also be interested in the development of the wind within that hour, i.e. is the wind speed increasing or decreasing, when do the highest or lowest wind speeds occur and what are the expected peaks. 
Furthermore, since wind power is proportional to the wind speed cubed, $P\propto ws^3$, higher resolution wind speed forecasts are necessary to more accurately obtain the expected wind power for a particular time period.
Bearing this in mind, this study decided to focus on multi-step forecasting with a 10-minute time resolution for all forecasts of 10-minutes, 1-hour and 4-hours ahead. 

The contributions of this paper can be summarised as:

\begin{itemize}
    \item We show the effectiveness of a generic framework for multi-step spatio-temporal forecasting, with GNNs to capture spatial correlations and optional update functions to learn temporal dependencies, such as a Transformer or LSTM network.
    \item Test and compare the performance of different Transformer architectures for use in wind forecasting, namely the vanilla Transformer, LogSparse Transformer, Informer and Autoformer. This is the first paper to formulate many of these in a GNN setting and the first to apply such models to wind forecasting. 
    \item We propose a new alteration of the Transformer architecture, namely the Fast Fourier Transformer (FFTransformer), and show its competitive performance in wind forecasting. The novel architecture is based on wavelet decomposition and an adapted Transformer architecture consisting of two streams. One stream analyses periodic components in the frequency domain with an adapted attention mechanism based on fast Fourier transform (FFT), and another stream similar to the vanilla Transformer, which learns trend components.
\end{itemize}

\section{Related Works}\label{sec:rw}
Autoregressive moving average methods, such as the autoregressive integrated moving average (ARIMA) model, are robust and easy to implement, making them popular for wind forecasting. 
Kavasseri \textit{et al.} \cite{kavasseri2009day} studied \textit{fractional}-ARIMA models to perform one- and two-day-ahead wind speed forecasts, managing to outperform both a persistence and an ARIMA model for four potential wind generation sites in North Dakota.
The persistence model is a commonly used benchmark in wind-speed forecasting, where the forecasted values, $\hat{ws}_{t+1}$ are simply taken as the last recorded value $ws_t$, i.e. $\hat{ws}_{t+1} = ws_t$. 
Since wind speed time series are characterised by both long-term trends and high-frequency variation, Singh \textit{et al.} \cite{singh2019repeated} proposed the RWT-ARIMA model, combining the ARIMA model with wavelet transform (WT) to decompose the signal into multiple sub-series with different frequency characteristics. 
Various decomposition techniques have been studied for wind time series, showing the potential benefits of introducing some signal decomposition into the forecasting models, such as complete ensemble empirical mode decomposition~\cite{da2021novel}, variational mode decomposition~\cite{wu2022interpretable} and wavelet packet decomposition~\cite{meng2016wind, liu2013forecasting}. 

Support vector regressor (SVR) and K-nearest neighbour (KNN) algorithms have also been popular within wind forecasting \cite{jorgensen2020wind, zendehboudi2018application, colak2012data}. 
The KNN-algorithm is based on finding similar points in the available data and can be fast in both training and testing. 
SVRs have been shown to yield very good forecasting performance \cite{zendehboudi2018application}, but do not scale well for larger datasets, resulting in longer computation times \cite{jorgensen2020wind}. 

In this study, we focus on neural network architectures, which lie at the heart of modern ML and have become increasingly popular for wind speed forecasting, due to their ability to model non-linear relationships. 
Multi-Layer Perceptrons (MLP) have been successfully used both in isolation \cite{sfetsos2002novel, more2003forecasting} and in combination with other methods \cite{shi2012evaluation, guo2011case}.
Recurrent (RNN) and convolutional neural networks (CNN) represent the quintessential DL architectures for sequence modelling and are typically favoured over MLPs for wind forecasting \cite{alkhayat2021review}.
The long short-term memory (LSTM) unit is an alteration of the vanilla RNN architecture, where gating mechanisms and skip connections are introduced to mitigate the problem of vanishing or exploding gradients \cite{schmidhuber1997long}. 
Li \textit{et al.} \cite{li2018multi} proposed a hybrid architecture, using empirical wavelet transform and the regularised extreme learning machine, together with an LSTM as the main predictor. 
Due to the recurrent architecture, the LSTM network relies on encoding all the relevant input information into a fixed-length memory cell, which can cause information loss and limit the network's ability to retain information across longer sequences. 
Within the context of natural language processing (NLP), the attention mechanism was introduced by Bahdanau \textit{et al.} \cite{bahdanau2014neural}, to allow the networks to directly attend to previous hidden states according to their importance.
Many studies have focused on integrating attention mechanisms with LSTMs to further help the models learn long-term dependencies that can improve forecasting performance \cite{li2020short, huang2021wind}.

Oord \textit{et al.} \cite{oord2016wavenet} proposed the WaveNet architecture for generating raw audio waveforms. 
The main ingredient of WaveNet is dilated causal convolution, which is a 1D convolutional operation where the causality ensures that the model cannot violate the sequence ordering, while the dilation increases the receptive field by skipping input values with a certain step. 
Dilated causal convolution is well suited for time series modelling and has been successfully deployed for some wind forecasting studies \cite{dong2021spatio, shivam2020multi}.
Other popular DL-based architectures used for wind forecasting also include deep belief networks \cite{wang2016deep}, RNNs with Gated Recurrent Units \cite{niu2020wind} and the N-BEATS model \cite{putz2021novel}.

Building on the attention mechanism, the Transformer was proposed by Vaswani \textit{et al.} \cite{vaswani2017attention}, as a new sequence transduction model, particularly focused on NLP. 
The Transformer is fundamentally different from previous models in that it does not rely on recurrence or convolution, making it better at learning long-term dependencies. 
However, since the complexity scales quadratically with sequence length, various alterations of the original architecture have been proposed to alleviate the computational limitations and make the models better suited for time series forecasting, such as the Longformer~\cite{beltagy2020longformer}, FEDformer \cite{zhou2022fedformer}, Temporal Fusion Transformer~\cite{lim2021temporal}, LogSparse Transformer~\cite{li2019enhancing}, Informer~\cite{zhou2021informer}, Reformer~\cite{kitaev2020reformer} and Autoformer~\cite{wu2021autoformer}. 
Both \cite{wang2022novel} and \cite{qu2022short} managed to outperform LSTM models for wind forecasting by using a Transformer, comprised of an encoder and decoder, as in \cite{vaswani2017attention}. 
A bidirectional LSTM-Transformer model achieved superior results compared to a gated recurrent unit (GRU) and an LSTM model in \cite{liu2021wind}.  
Wang \textit{et al.} \cite{wang2022hybrid} proposed a model based on the Informer together with convolutional layers that extract temporal features at different frequencies, to forecast the average wind power over the next three hours. 
The Spatial-Temporal Graph Transformer Network (STGTN) extends the previous research by leveraging both spatial and temporal correlations within a wind farm, to more accurately forecast wind speeds at a turbine level, 10 minutes - 1 hour ahead~\cite{pan2022short}. 
Despite some efforts at employing different Transformer-based architectures for wind forecasting, namely the vanilla Transformer and Informer, the research have nevertheless been relatively scarce. 
In this study, we therefore aim to further research the performance of different Transformer architectures, investigating the Informer, Autoformer and LogSparse Transformer, which are yet to be thoroughly tested for wind forecasting.

\begin{figure}
    \centering
    \includegraphics[width=6cm]{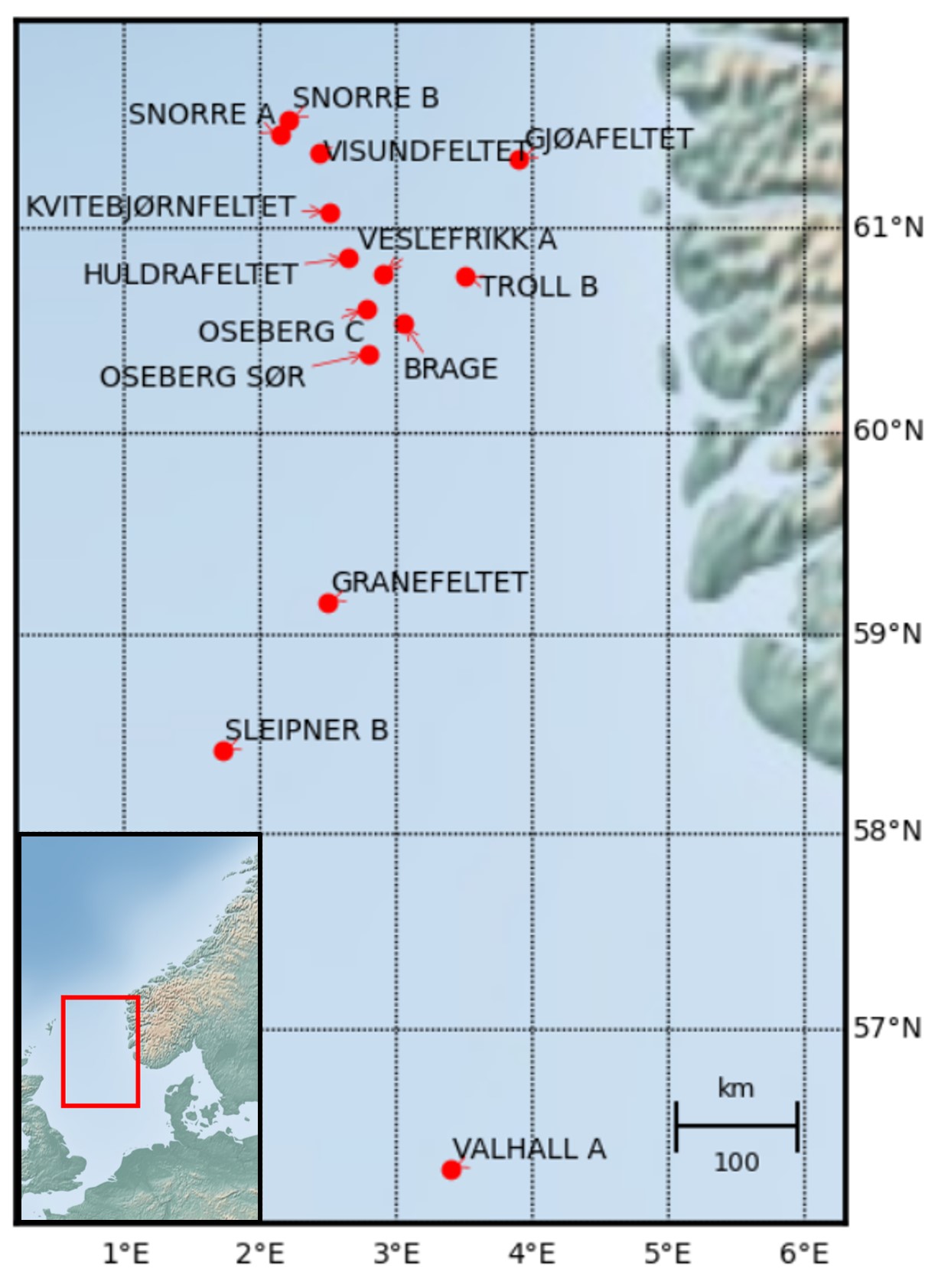}
    \caption{The 14 measurement stations in the North Sea used to construct the dataset.}
    \label{fig:stations}
\end{figure}

Spatial dependencies can be important for improving meteorological forecasts, such as for wind.
Considering the 14 measurement stations depicted in Fig.~\ref{fig:stations} used for this study, spatio-temporal forecasting aims to leverage spatial dependencies between different stations, through the physical distance between them, to improve the local forecasts for the different sites.
The spatio-temporal interdependence of different physical locations can be particularly important for meteorological forecasts since physical properties such as wind fields, are non-stationary in both space and time, meaning that historical time series for different physical locations should be jointly considered to better learn global trends and propagation in both space and time. 
Some studies organise spatial data, i.e. the physical location of different measurement points, into an ordered grid, where the features for a particular location are assigned to a specific cell~\cite{hu2019very, zhu2019learning, wang2021review}. 
A CNN is then used to extract spatial features, while another network, such as a CNN~\cite{hu2019very} or LSTM~\cite{zhu2019learning}, is used to learn temporal correlations. 
However, considering the complex topology of the different measurement stations in Fig.~\ref{fig:stations}, the strict ordering of the input data required for CNNs might not be able to effectively represent the underlying spatial relationships. 
Graph neural networks (GNN) can better facilitate arbitrary spatial ordering and have therefore been popularly used for spatio-temporal forecasting.
Khodayar \textit{et al.} \cite{khodayar2018spatio} made forecasts at different wind sites simultaneously, where each site was represented by a node in an undirected graph, using a Graph Convolutional Network (GCN) and an LSTM to extract spatial and temporal features, respectively. 
Similarly, Sta\'{n}czyk \textit{et al.} \cite{stanczyk2021deep} also used a GCN, but with a CNN to learn temporal dependencies. 
Instead of using static edge features, Wang \textit{et al.} \cite{wang2022dynamic} constructed edge features based on the time-varying spatial correlation between wind sites and used a GCN for wind farm cluster power forecasting. 
Finally, the M2STAN model was also proposed for spatio-temporal multi-step wind power forecasting~\cite{wang2022m2stan}, which employs a Graph Attention Network (GAT) and a Bidirectional GRU for spatial and temporal correlation modelling, respectively, along with a Transformer network for multi-modal feature fusion.
This study will further build on the ideas of using GNNs for spatio-temporal forecasting, but also focus on recent advancements in Transformer-based architectures for time series analysis. 
Furthermore, the FFTransformer is also proposed, which is a novel Transformer architecture for time series forecasting, based on signal decomposition and learning trend and periodic components separately. 
Finally, even though many studies focus on single-step forecasts, predicting average wind speeds over some future interval, we also aim to improve predictions by considering multi-step forecasting, producing higher resolution time series for the forecasting windows.

\section{Theory}\label{sec:theo}
\begin{figure*}[b]
\begin{minipage}[b]{0.38\linewidth}
    \centering
    \includegraphics[width=6cm]{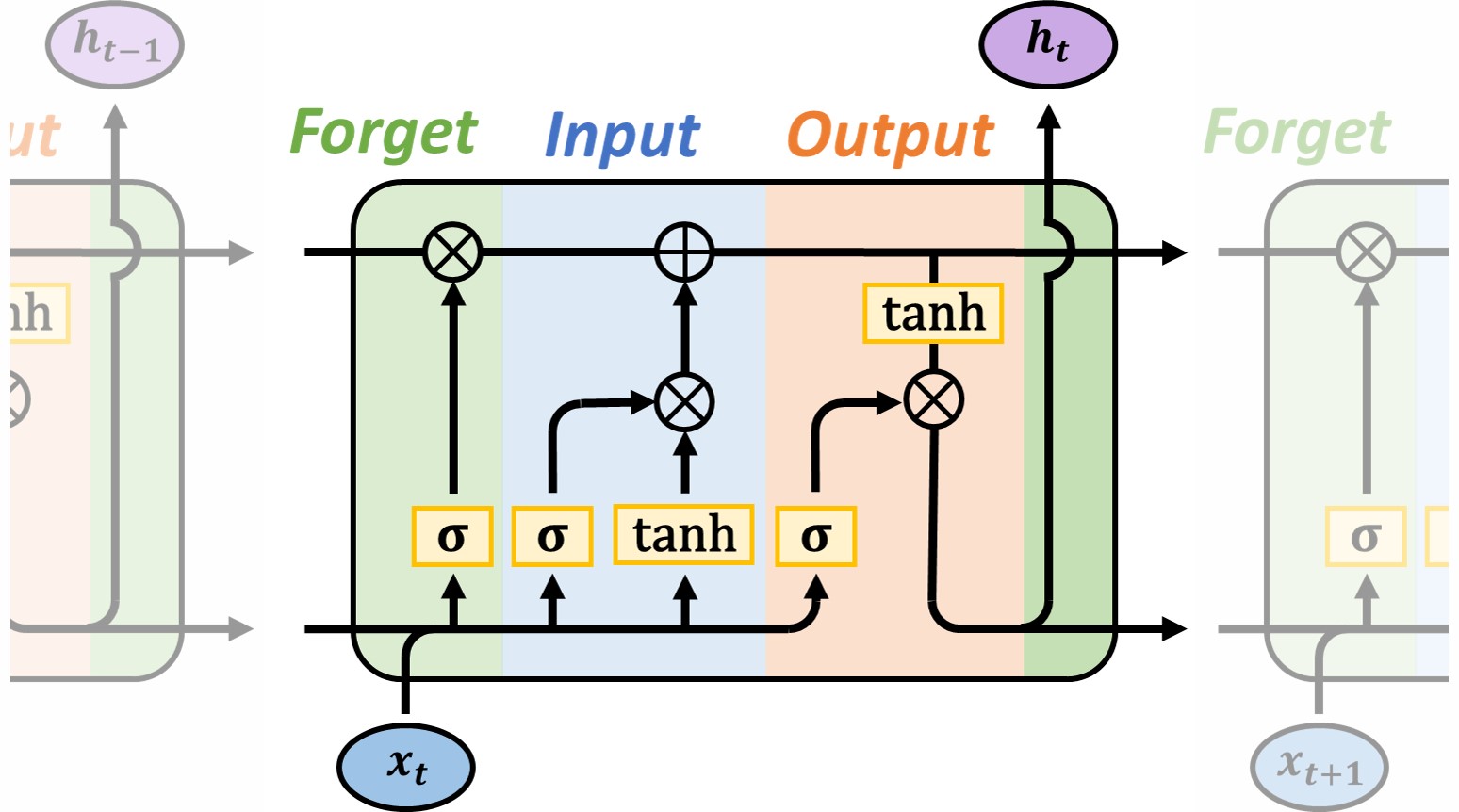}
    \caption{Visualisation of an RNN with LSTM units having forget, input and output gating mechanisms.}
    \label{fig:lstm}
    \vspace{\baselineskip}
    \includegraphics[width=6cm]{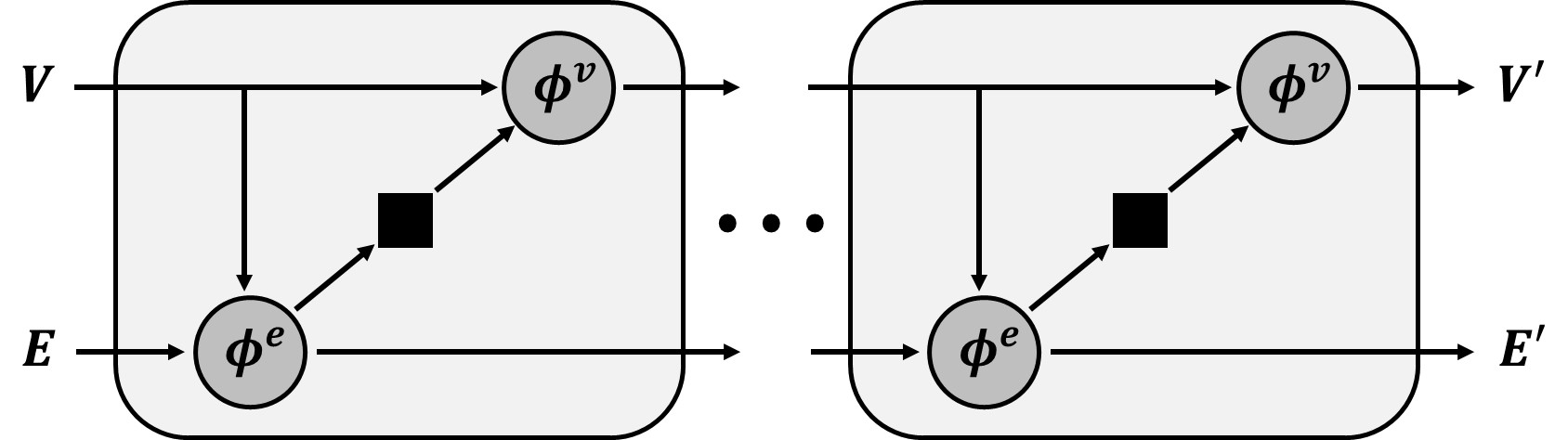}
    \caption{Visualisation of the GNN architecture for graphs with edge and node features. $\phi^{(\cdot)}$ and \img{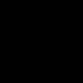} represent the update functions and edge-to-node aggregation, respectively.}
    \label{fig:gnn}
\end{minipage}%
    \hfill%
\begin{minipage}[b]{0.58\linewidth}
    % \centering
    % \includegraphics[width=7.8cm]{Figures/GNN.jpg}
    % \caption{Visualisation of the stacked GNN architecture for graphs with edge and node features. $\phi^{(\cdot)}$ and \img{Figures/agg_oper.jpg} represent the update functions and edge-to-node aggregation, respectively.}
    % \label{fig:gnn}
    \centering
    \includegraphics[width=9.5cm]{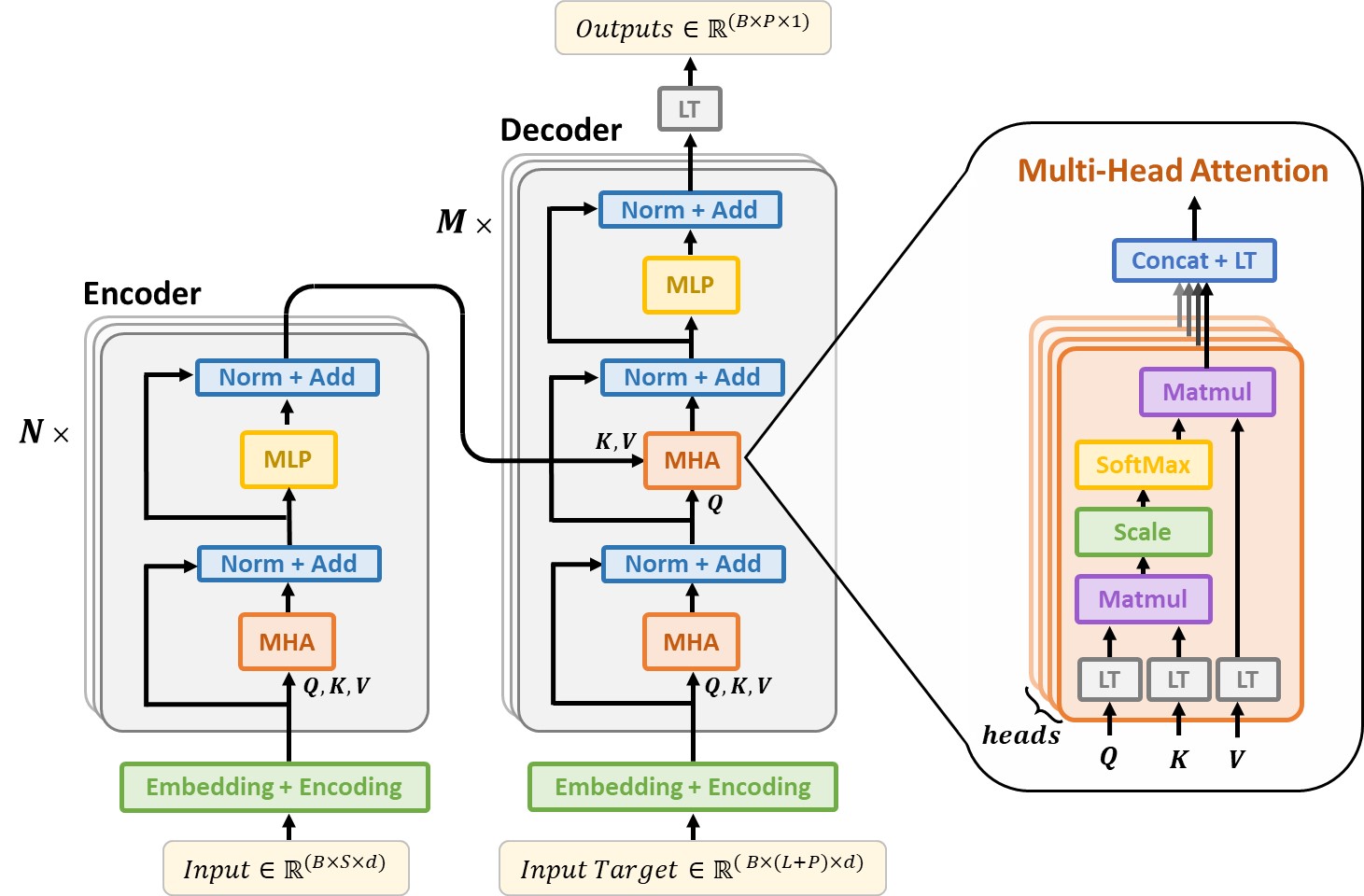}
    \caption{The Transformer architecture \cite{vaswani2017attention} in an encoder-decoder setting, adapted to facilitate forecasting rather than categorical outputs, with multivariate inputs and univariate outputs. $B$ in the inputs and outputs refer to the batch size, for a single prediction $B = 1$.}
    \label{fig:trans}
\end{minipage} 
\end{figure*}

\subsection{Multi-Layer Perceptron}\label{sec:theo:mlp}
MLPs are feed-forward networks that learn weights, $\theta$, which map the input to output, $y \approx f(x|\theta)$.
A chain structure, where multiple layers are stacked, gives depth to the model, as $\hat{y} = f^{(3)}(f^{(2)}(f^{(1)}(x|\theta_1)|\theta_2)|\theta_3)$, for a model with two hidden layers.
To improve the learning ability of the model, non-linearities such as the ReLU or sigmoid functions, are applied to the neuron outputs. 
Optimal weights are determined by minimising a differentiable loss function, using backpropagation, which will update network weights by propagating gradients of the weights with respect to the loss function, back through the network \cite{goodfellow2016deep}.

\subsection{LSTM}\label{sec:theo:lstm}
Even though sequence analysis using DL has largely been dominated by RNNs, the original architecture is prone to exploding or vanishing gradients, resulting in significant information loss for longer sequences. 
The long short-term memory (LSTM) unit introduces gating mechanisms and skip connections to alleviate some of these shortcomings \cite{schmidhuber1997long}.
An illustration showing the internal workings of an LSTM unit is given in Fig.~\ref{fig:lstm}, where an input gate controls the contribution of the new input, $x_t$, to the memory, while the forget and output gates control what information to be kept in memory and encoded in the output, $h_t$, respectively.

\subsection{Transformer Architectures}\label{sec:theo:tran}
The Transformer architecture was proposed as a model for sequence analysis, without any recurrence or convolution, but instead based on the attention mechanism \cite{vaswani2017attention}. 
If we consider a time series forecasting task with $d$ input features to predict a single output feature, the Transformer model should take an input, $x^{(e)} \in \mathbb{R}^{S \times d}$, and produce an output, $y \in \mathbb{R}^{P \times 1}$, where, $S$, is the look-back window and, $P$, the forecasting horizon. 
The original architecture consists of an encoder and a decoder, as shown in Fig.~\ref{fig:trans}, where the encoder should encode a longer input into a hidden state representation for the decoder to decode. 
Inputs to the decoder, $x^{(d)} \in \mathbb{R}^{(L+P) \times d}$, are typically shorter than those to the encoder, containing the last $L$ elements of the encoder inputs, where $L < S$, and some placeholders are used in place for the last $P$-indices, which correspond to the forecasting locations. 

Inputs are first linearly transformed to $E$-dimensional space and then added with some positional encoding, so that the model can make use of the sequence order, without any recurrence.
Unless otherwise is stated, it will be assumed that the positional encoding added to the encoder and decoder inputs will be the sine-cosine positional encoding proposed in the original architecture \cite{vaswani2017attention}. 
The multi-head attention (MHA) block in the encoder employs full self-attention, where each attention operation can attend to the full input sequence. 
Scaled dot-product attention takes $Q$, $K$ and $V$ as inputs, which represent the queries, keys, and values, respectively.
Dot products between keys and queries are computed, before being passed to a softmax function to obtain the attention weights, which are then multiplied by the values to produce the final outputs. 
This process can be summarised as 
\begin{equation}
    \resizebox{0.9\hsize}{!}{$\text{Attn}(Q, K, V) = \text{softmax}(\frac{(QW^Q)(KW^K)^T}{\sqrt{d_k}})(VW^V),$}
\end{equation}
where $W^{(\cdot)} \in \mathbb{R}^{E\times d_{(\cdot)}}$ are weights of the different linear transformations (LT), $E$ is the hidden dimensionality and, $d_k$, the dimension of keys and queries. 
Performing multi-head attention, with $h$ separate projections, allows the module to simultaneously attend to different information in the input series. 
Outputs from the MHA are then concatenated and linearly projected to produce a single output. 

A residual connection and layer normalisation are applied to the outputs before the signal is passed to an MLP, typically with a single hidden layer, which is applied to each position in the series separately.
Multiple encoder layers, with different weights, are stacked to add depth to the model and hence a stronger function approximation ability. 

The decoder follows almost the same architecture as the encoder, but with an additional MHA block, referred to as the cross-attention, which precedes the MLP.
The cross-attention module is the same as the other MHA blocks but takes the encoder outputs as inputs to the keys and values.
Additionally, the first MHA block in the decoder use masking to prevent information flow from subsequent positions.

Even though the attention mechanism alleviates the problem of information loss for longer sequences, by allowing every position to directly attend to all other positions, its complexity scales quadratically with sequence length. 
Furthermore, since the Transformer was originally developed for machine translation and other NLP tasks, various modifications have emerged, which aim to both combat the complexity limitations and further adapt the architecture to facilitate time series forecasting problems. 
A short summary of some of these will now be given in the subsequent sections.

\subsubsection{LogSparse Transformer \cite{li2019enhancing}} \label{sec:theo:tran:logsp}
The LogSparse Transformer introduces two novel alterations. 
Sparse attention, where each position is only allowed to attend to other positions with an exponential step size and itself, significantly reduces the space complexity. % of the Transformer. 
Since the point-wise attention operation described in Sec. \ref{sec:theo:tran} is insensitive to local context, causal 1D-convolution was used to compute keys and queries, instead of point-wise linear transformations. 
The modified transformation of keys and queries, which will here be referred to as convolutional attention, might be particularly advantageous for time series forecasting, as local context could be very important for signals characterised by high-frequency fluctuations or noise.

\subsubsection{LogSparse Transformer \cite{li2019enhancing}} \label{sec:theo:tran:logsp}
The LogSparse Transformer introduces two novel alterations. 
Sparse attention, where each position is only allowed to attend to other positions with an exponential step size and itself, significantly reduces the space complexity.
Since the point-wise attention operation described in Sec. \ref{sec:theo:tran} is insensitive to local context, causal 1D-convolution was used to compute keys and queries, instead of point-wise linear transformations. 
The modified transformation of keys and queries, which will here be referred to as convolutional attention, might be particularly advantageous for time series forecasting, as local context could be very important for signals characterised by high-frequency fluctuations or noise.

\subsubsection{Informer \cite{zhou2021informer}} \label{sec:theo:tran:infor}
Instead of introducing sparsity through a fixed pattern decided by heuristic methods, the \textit{ProbSparse} self-attention mechanism proposed for the Informer introduces sparsity by locating the most dominant queries and only allows keys to attend to these.
Dominant queries are taken as those that maximise a surrogate for the KL-divergence between a uniform distribution and the query's attention probability distribution. 
Furthermore, self-attention distilling in the encoder highlights dominant attention by halving cascading layer inputs through 1D-convolution and MaxPooling, which makes the model much more efficient for very long sequences. 

\begin{figure*}[b]
    \begin{subfigure}{0.24\textwidth}
        \includegraphics[height=3.6cm]{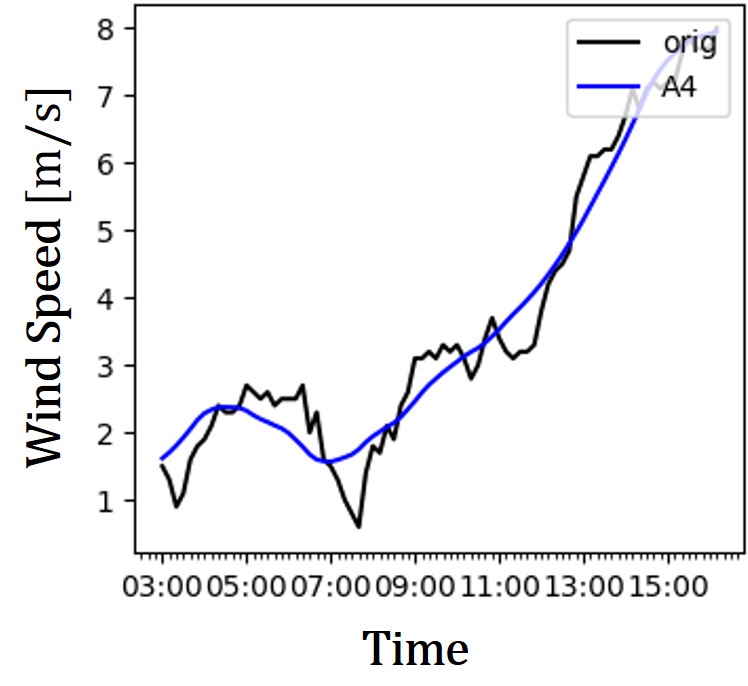}
        \caption{Approximation and Input} \label{fig:wt_a}
    \end{subfigure}
    \hspace*{\fill}
    \begin{subfigure}{0.24\textwidth}
        \includegraphics[height=3.6cm]{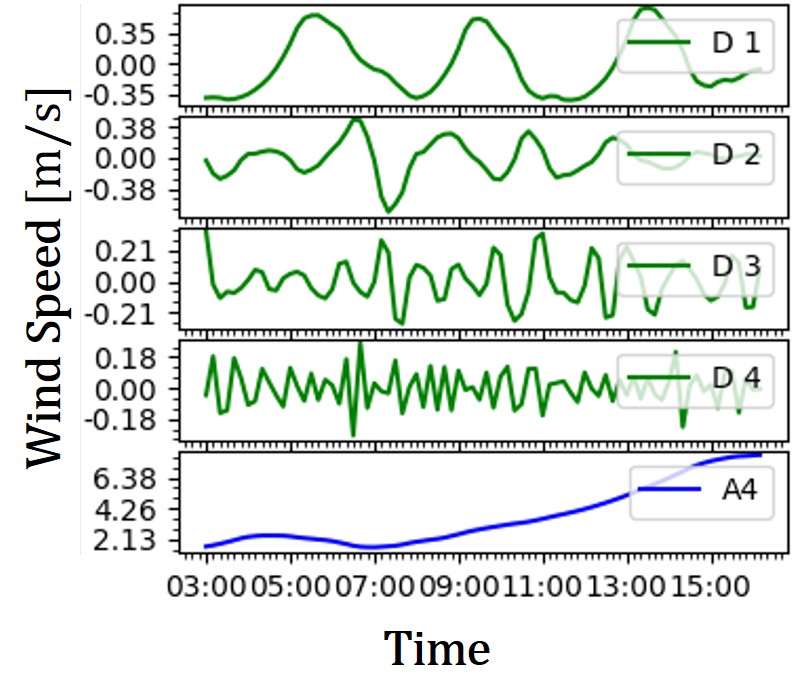}
        \caption{Inverse MDWD} \label{fig:wt_b}
    \end{subfigure}
    \hspace*{\fill}
    \begin{subfigure}{0.24\textwidth}
        \includegraphics[height=3.6cm]{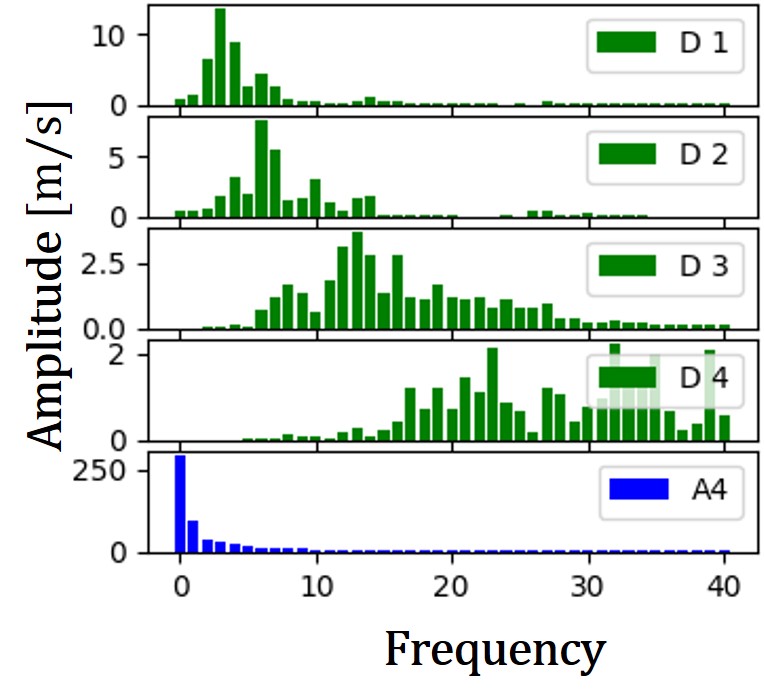}
        \caption{ABS(FFT(Inverse MDWD))} \label{fig:wt_c}
    \end{subfigure}
    \hspace*{\fill}
    \begin{subfigure}{0.24\textwidth}
        \includegraphics[height=3.6cm]{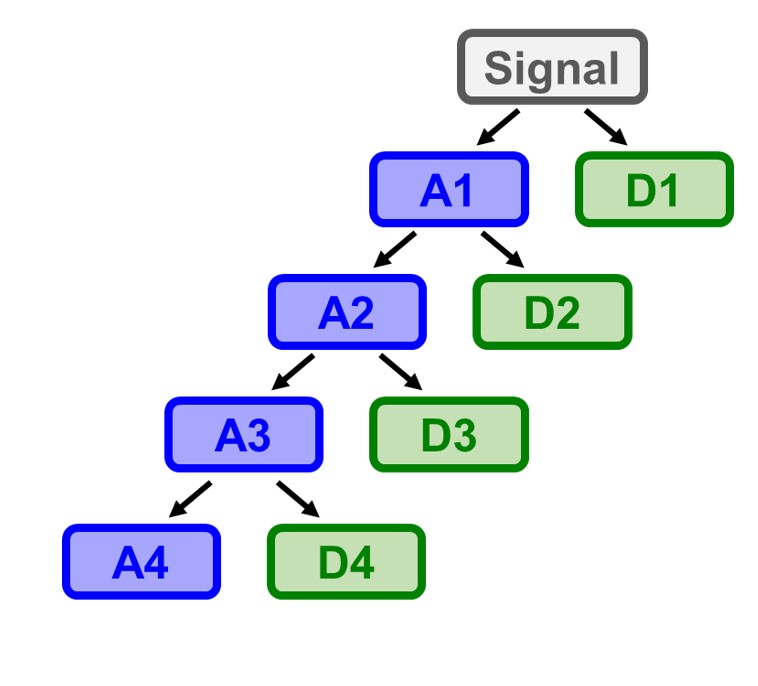}
        \caption{MDWD Architecture} \label{fig:wt_d}
    \end{subfigure}
\caption{MDWD applied to a wind speed time series. The FFT outputs clearly exhibit the different frequency characteristics of the sub-components and plotting the approximation together with the input signal visualises how this component retains the trend information.}
\end{figure*}

\subsubsection{Autoformer \cite{wu2021autoformer}} \label{sec:theo:tran:auto}
Unlike the Informer and LogSparse Transformer, Wu \textit{et al.} \cite{wu2021autoformer} proposed significant alterations to both the overall Transformer architecture and the attention module, to better facilitate time series forecasting. 
Instead of the scaled dot-product attention, the Autoformer introduces the Auto-Correlation mechanism, which uses keys and queries to decide on the most important time-delay similarities through autocorrelation and a time-delay aggregation, which rolls the series according to the selected time delays, before adding them together to produce the outputs. 
Different to point-wise attention, the Auto-Correlation mechanism finds dependencies based on periodicity and is specifically designed for time series forecasting.
Series decomposition is applied after every Auto-Correlation and MLP module, using average pooling to decompose a signal, $X$, into trend and seasonal components, $X_t$ and $X_s$, respectively: 
\begin{align}
    X_t &= \text{AvgPool}(\text{Padding}(X)) \\
    X_s &= X - X_t .
\end{align}\label{eq:seriesdecomp}

\subsection{Graph Neural Networks}\label{sec:theo:gnn}
A graph can be defined as a tuple with node- and edge-specific features, $G=(V, E)$ \cite{battaglia2018relational}.
For the spatial forecasting problem, node features, $v_i \in V$, can represent attributes associated with a particular measurement station, while edge features, $e_{ij} \in E$, contain information on the relationship between two nodes, $i$ and $j$, such as the Euclidean distance between two stations. 
A GNN is comprised of stacked graph blocks, which perform per-edge and per-node updates in the following order: 
\begin{align}
    e'_{ij} &= \phi^e(e_{ij}, v_i, v_j) \label{eq:gnn1} \\
    v'_j &= \phi^v(v_j, \bar{e}'_j) \label{eq:gnn2} ,
\end{align}
where $\phi^{(\cdot)}$ are the update functions, which could be represented by a neural network such as an LSTM, MLP or Transformer. 
$(\cdot)'$ and $\bar{(\cdot)}$ represent updated and aggregated features, respectively.
Aggregated edge features, $\bar{e}'_j$, are computed using an aggregation function, $\rho^{e \to v}$, as
\begin{equation}
    \bar{e}'_j = \rho^{e \to v}(E'_j), \text{   where  } E'_j = \{e'_{ij}|\forall i \in \mathcal{R}_j\},
\end{equation}
which could for example be a sum or mean operation.
$\mathcal{R}_j$ is an index set containing all nodes sending to node $j$, and defines the connectivity of the graph. 
A visualisation of the GNN architecture is given in Fig.~\ref{fig:gnn}.

\subsection{Multilevel Wavelet Decomposition}\label{sec:theo:wt}
Discrete wavelet decomposition (DWD) is a method which can be used to decompose a time series signal into its trend and periodic components, referred to as the approximate and detail signals, respectively, using low- and high-pass filters. 
An input signal $x={x_1, x_2, ..., x_S}$ is decomposed by convolving a low pass filter $l = {l_1, l_2, ..., l_K}$ and a high pass filter $h = {h_1, h_2, ..., h_K}$, where $K \ll S$, over the input:
\begin{align}
    \text{A1}_i &= \sum_{k=1}^K x_{2i+k-1} \cdot l_k \label{eq:wt_1} \\
    \text{D1}_i &= \sum_{k=1}^K x_{2i+k-1} \cdot h_k, \label{eq:wt_2}
\end{align}
where $\text{A1}_i$ and $\text{D1}_i$ are the $i^\text{th}$ elements of the approximate and detail components obtained from a single level discrete wavelet decomposition. 
DWD can also be stacked in multiple layers, referred to as multilevel DWD (MDWD), to extract multiple periodic components of different frequency characteristics. 
In a multi-layer setting, the approximate component from one layer is fed as input to the next, resulting in multiple detail and a single approximate signal, as shown in Fig~\ref{fig:wt_d}, where `D1', `D2', `D3' and `D4' are the detail components and `A4' the approximation.
The Daubechies wavelet, Db4, for the low- and high-pass coefficients, $l$ and $h$ in eq. (\ref{eq:wt_1}) and (\ref{eq:wt_2}), has been shown to be suitable for wind forecasting \cite{catalao2011short}.
The decomposition of a wind speed time series is shown in Fig.~\ref{fig:wt_b}, where inverse MDWD is applied independently to the outputs from the MDWD.
By applying FFT to each time series in Fig~\ref{fig:wt_b}, it can be seen in Fig.~\ref{fig:wt_c} that detail (D1, D2, D3, D4) and approximate (A4) components yield very different frequency characteristics.
The four detail components exhibited clear periodic information, with peaks at different frequencies, while the approximation, `A4', contained most of the low-frequency trend information, clearly visualised in Fig.~\ref{fig:wt_a}.
As a result, we argue that MDWD is a suitable decomposition method for wind speed time series, where the special filters and multiple layers potentially allow us to extract more informative trend and periodic information than simpler decomposition methods, such as series decomposition. 

\section{Methodology}\label{sec:meth}

\subsection{Dataset}\label{sec:meth:data}
Due to the future potential for offshore wind energy \cite{diaz2020review}, this  study decided to focus on offshore wind speed forecasting, using meteorological measurements recorded on the Norwegian continental shelf.
The data was made available by the Norwegian Meteorological Institute and is openly available through the Frost API\footnote{\url{https://frost.met.no/index.html}}.
Ten-minute averaged measurements on air temperature, air pressure, dew point, relative humidity, wind direction and speed and maximum wind speed in the 10-minute interval were used as input features to forecast the wind speed time series.
Wind direction was decomposed into its sine and cosine components in order to fully capture the circular characteristics.
For every time-step, we would therefore have a feature matrix, $f_t \in \mathbb{R}^{N\times8}$, corresponding to the eight recorded measurements for $N$ available stations. 
The forecasting problem then becomes, $\hat{V}_{(t+1)}, \hat{V}_{(t+2)}, ..., \hat{V}_{(t+P)} = F(f_{(t)}, f_{(t-1)}, ..., f_{(t-(S-1))})$, where $F$ is the model, $P$ the number of future time-steps to forecast, $\hat{V}_{(t+j)} \in \mathbb{R}^{N\times1}$ the predicted wind speeds at time, $(t+j) \forall j\in (1, 2, ... P)$, and $S$ is the look-back window, i.e. the number of previous time-steps used to make the forecasts.
For the period between June 23, 2015, and February 28, 2022, 14 out of 25 available stations had some periods with measurements on all the relevant meteorological variables, and are shown in Fig.~\ref{fig:stations}.

\begin{figure*}
    \centering
    \includegraphics[width=12.6cm]{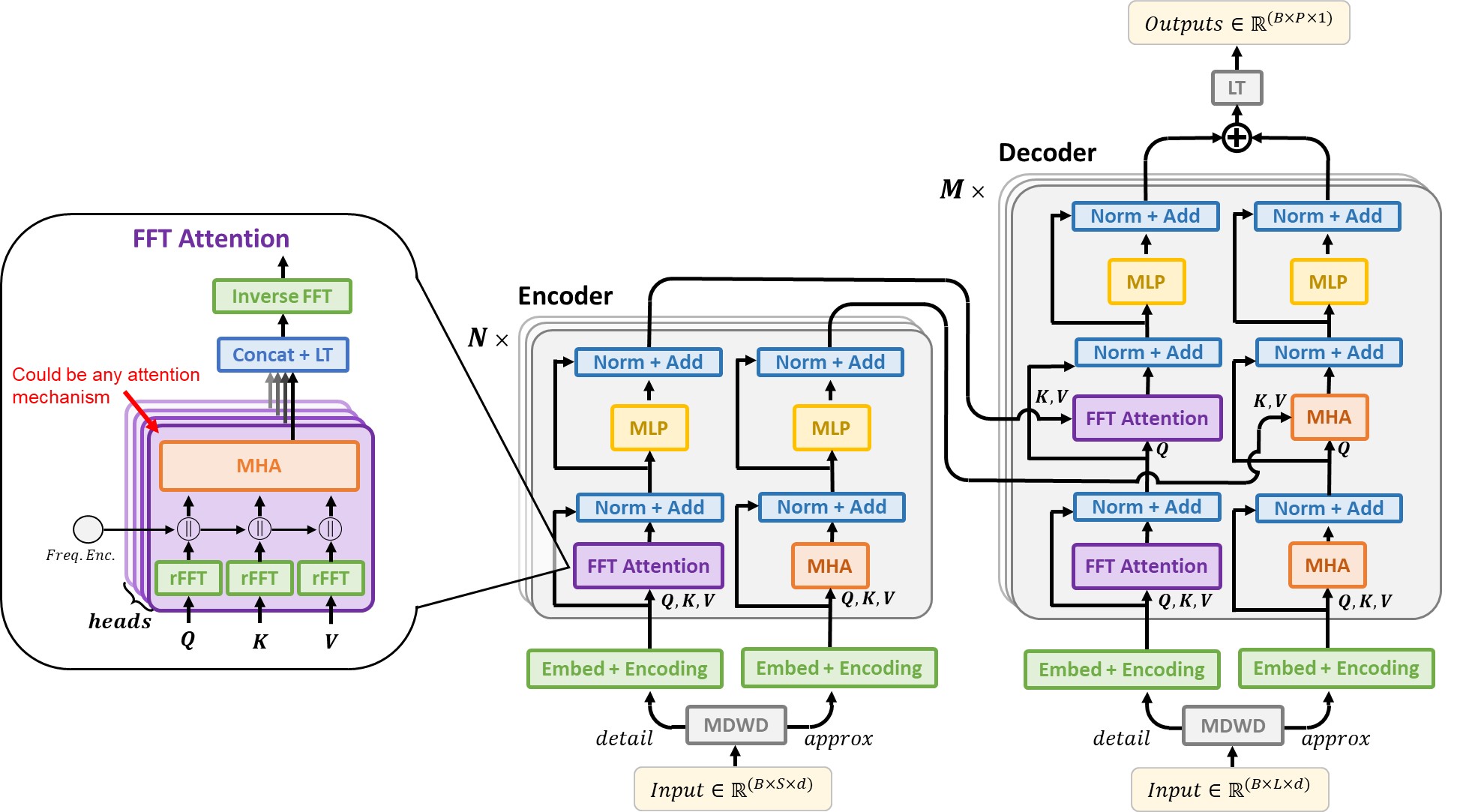}
    \caption{FFTransformer in an encoder-decoder setting. Two streams focus on trend and periodic signal components separately. The FFT-Attention is employed for the periodic analysis in the frequency domain, using an MHA mechanism together with FFT.}
    \label{fig:fftrans}
\end{figure*}

The first 60\% of the data was used for training, whereas the following 20\% and the remainder were used for validation and testing, respectively. 
If measurements for a single time-step were missing for a particular station, linear interpolation was used to fill the missing entries. 
However, if there were consecutive time-steps missing, the station would not be considered for these periods.
For different periods, there would be a variable number of stations that had available data, meaning that the models should be able to take any subset of the 14 stations as input.

\subsection{Fast Fourier Transformer}\label{sec:meth:fft}

Since many time series, such as wind measurements, are characterised by both trend and periodic components, we propose the Fast Fourier Transformer (FFTransformer), based on signal decomposition and an adapted attention mechanism, named FFT-Attention. 
Given the different frequency characteristics for wind speed time series discussed in Sec.~\ref{sec:theo:wt}, the FFTransformer is comprised of two streams, one which analyses signals with clear periodicity and another that should learn trend components, i.e. the detail and approximate signals in Fig.~\ref{fig:wt_b}, respectively. 
Here, we used MDWD to decompose the input signals into its periodic and trend components, due to its capability to clearly extract different level frequencies using multiple levels of low- and high-pass filters. 
Nevertheless, the architecture discussed is not limited to MDWD and other methods for signal decomposition could be used, such as the `Series Decomposition' in the Autformer. 

As shown in Fig.~\ref{fig:fftrans}, the right-hand stream in both the encoder and decoder were exactly the same as for the original Transformer in Fig.~\ref{fig:trans}. 
However, to better facilitate the analysis of periodic signals in the left-hand stream, we introduce the FFT-Attention mechanism, which performs attention in the frequency domain. 
In particular, as the first operation of FFT-Attention, FFT is applied to the key, query and value inputs.
Real and imaginary components of the FFT outputs are concatenated with the frequency values, to provide information on the corresponding frequency for the values in a particular position, similar to the motivation behind positional encoding, before being fed to an MHA block.
Outputs from the FFT-Attention module are then concatenated and projected, as for the other attention mechanisms, and finally, inverse FFT transform values back to the time domain. 
Any attention mechanism could be used in place of all the MHA blocks in Fig.~\ref{fig:fftrans}, such as the \textit{ProbSparse} or convolutional attention. 
The final network outputs from both streams (FFT Attention and Trend) are at last added and linearly transformed to produce the predictions. 

After some experimentation,  it was found that not adding temporal or positional encoding to the detail components (i.e. inputs to the FFT-Attention) yielded the best results.
This seemed sensible, as adding positional encoding in the time domain would not translate to the frequency domain.
Instead, the concatenation of frequency values to the FFT outputs, served somewhat the same purpose as the positional encoding did for the trend components. 
Nevertheless, since this method was fairly trivial, further research studying better encoding strategies for the frequency domain could be desirable.

\subsection{Spatio-Temporal Framework}\label{sec:meth:spatemp}

All models were constructed in an encoder-only setting, i.e. without decoders, and used as update functions, $\phi^{(\cdot)}$, in two-layer GNNs. 
To facilitate the GNN framework, the dataset was constructed as graphs. 
Measurement stations were represented by nodes, with the historical time series of the eight meteorological variables assigned to the input node features.
Since there would be a variable number of available measurement stations at different times, the input graphs would not be the same at different times, but with a variable number of nodes. 
This was desirable, as it meant that we did not have to discard the data for all stations or interpolate missing values for a particular time interval where a few stations had missing data, as would be the typical case for a CNN. 
Considering a node, $i$, its input features were $v_i \in \mathbb{R}^{1 \times (S+P) \times 8}$, where $S$ is the historical look-back window, i.e. the number of previous time-steps used to predict the next $P$ time-steps.
Placeholders were used for the last $P$ indices of the node input features to the first graph block, since we do not have information on recorded values in the future.
These values were set to either the last available measurements, meaning the values at position $S$, or the mean of the input series, i.e. the mean of each feature over positions $(1, 2, ..., S)$.
Zero-values were used as placeholders for the seasonal inputs to the Autoformer, as well as for the detail components in the FFTransformer model. 
It should be noted that placeholders were added subsequent to the `MDWD' and prior to the `Embed + Encoding' block in Fig.~\ref{fig:fftrans}, meaning that inputs to the MDWD were $v_i \in \mathbb{R}^{1 \times S \times 8}$. 
Differences in latitude and longitude between stations were assigned to the input edge features as $e_{ij} = [(\text{lon}_j - \text{lon}_i), (\text{lat}_j - \text{lat}_i)]$, for two stations, $i$ and $j$.
A pseudocode summarising the Spatio-Temporal forecasting framework is given in Algorithm~\ref{alg:pseudo}.

\subsection{Experimental Set-up}\label{sec:meth:setup}
\begin{algorithm*}
\hspace*{\algorithmicindent} \textbf{Input:} \\
    \hspace*{\algorithmicindent} \hspace{0.3cm} $V\in \mathbb{R}^{N \times S \times 8} $ \Comment{recorded values for the 8 features for previous $S$ times} \\
    \hspace*{\algorithmicindent} \hspace{0.3cm} $T\in \mathbb{R}^{1 \times (S+P) \times 4}$ \Comment{time stamps for previous ($S$) and future ($P$) times, 4 $\to$ \{minute, hour, day, week\}} \\
    \hspace*{\algorithmicindent} \hspace{0.3cm} $E\in \mathbb{R}^{M\times 2} $ \Comment{distance in latitude and longitude between neighbouring stations} \\
    \hspace*{\algorithmicindent} \hspace{0.3cm} $\mathcal{R} \in \mathbb{Z}^{M\times 2} $ \Comment{index set containing sender and receiver indices} \\
    \hspace*{\algorithmicindent} \hspace{0.3cm} $\phi^{v}$ \Comment{node update function, e.g. FFTransformer or LSTM} \\
    \hspace*{\algorithmicindent} \hspace{0.3cm} $\phi^{e}$ \Comment{edge update function} \\
\hspace*{\algorithmicindent} \textbf{Output:} \\
    \hspace*{\algorithmicindent} \hspace{0.3cm} $\hat{V} \in \mathbb{R}^{N \times P \times 1} $ \Comment{predicted wind speeds for future $P$ time steps}
    \hrule
  \begin{algorithmic}[1]
    \If{$\phi^v$ is FFTransformer}
        \For{$d_i = 1, ..., 8$}
            \State $V_{d_i}^{(A1)}, V_{d_i}^{(D1)}, V_{d_i}^{(D2)}, V_{d_i}^{(D3)}, V_{d_i}^{(D4)} \gets \text{MDWD}(V[:, :, d_i])$     \Comment{for a four-layer MDWD (see Sec.~\ref{sec:theo:wt})}
        \EndFor
        \State $V^{trend} \gets \text{concat}(V^{(A1)}_{1:8})$
        \State $V^{freq} \gets \text{concat}(V^{(D1:D4)}_{1:8})$
        \State $V^{trend} \gets \text{concat}(V^{trend}, V^{trend}[:, S, :].\text{repeat}(1, P, 1))$   \Comment{persistence placeholders to forecast locations}
        \State $V^{freq} \gets \text{concat}(V^{freq}, \text{zeros}(N, P, 8))$   \Comment{zero-value placeholders to forecast locations}
        \State $V^{freq} \gets NodeEmbed^{freq}(V^{freq})$      \Comment{learned linear projection as node and temporal embedding}
        \State $V^{trend} \gets NodeEmbed^{trend}(V^{trend}) + TempEmbed(T) + PosEmbed(S+P)$    \Comment{positional embedding as in \cite{vaswani2017attention}}
        \State $V^{(0)} \gets [V^{freq}, V^{trend}]$
    \Else
        \State $V \gets \text{concat}(V, V[:, S, :].repeat(1, P, 1))$   \Comment{persistence placeholders to forecast locations}
        \State $V^{(0)} \gets NodeEmbed(V) + TempEmbed(T) + PosEmbed(S + P)$    \Comment{PosEmbed not used for ST-MLP/LSTM}
    \EndIf
    \State $E \gets E.repeat(1, S + P, 1)$      \Comment{extend edge features to series length}
    \State $E^{(0)} \gets EdgeEmbed(E) + TempEmbed(T) + PosEmbed(S + P)$    \Comment{PosEmbed not used for ST-MLP/LSTM}
    \For{$l = 1, ..., L$}     \Comment{for $L$ number of  stacked graph blocks}
        \State $E^{(l)} \gets \phi^e (E^{(l-1)}, V^{(l-1)})$    \Comment{update edge features between stations}
        \State $\overline{e_j^{(l)}} \gets \rho^{e\to v}(e_{ij}^{(l)}) \hspace{0.25cm} \forall i \in \mathcal{R}_j$ \hspace{0.25cm} \textbf{for all} $j\in (1, 2, ..., N)$ \Comment{aggregate edge features, assumed to be mean}
        \State $V^{(l)} \gets \phi^v (V^{(l - 1)}, \overline{E^{(l)}})$     \Comment{update node features for each station}
    \EndFor
    \State \textbf{Return:} $\hat{V} \gets V^{(L)}[:, -P:, :]$ 
    
 \caption{Pseudocode for the spatio-temporal forecasting framework. Slicing of tensors follow a Python syntax.}\label{alg:pseudo}
\end{algorithmic}
\end{algorithm*}

\begin{table*}[]
    \small
    \centering
    \captionsetup{justification=raggedright,singlelinecheck=false}
    \caption{Final model parameters after tuning}
    \resizebox{\textwidth}{!}{%
    \begin{tabular}{lcccccccl}
        \hline
         Parameter                   & ST-MLP           & ST-LSTM           & ST-Transformer    & ST-LogSparse      & ST-Informer       & ST-Autoformer     & ST-FFTransformer  & Explanation\\
         \hline
         $d$	                     & (64, 64, 32)$^*$	& (16, 32, 64)$^*$  & (64, 64, 32)$^*$  & (64, 64, 32)$^*$  & (64, 64, 32)$^*$  & (64, 64, 32)$^*$  & (64, 64, 32)$^*$  & Input/Output dimension for every GNN layer \\ 
         $d_{\text{hidden}}$            & (256, 256, 128)$^*$ & (64, 128, 256)$^*$	& (256, 256, 128)$^*$ & (256, 256, 128)$^*$ & (256, 256, 128)$^*$ & (256, 256, 128)$^*$ & (256, 256, 128)$^*$ & Hidden dimensionality of MLP modules \\
         GNN Layers 	             & 2	            & 2             	& 2	                & 2	                & 2	                & 2	                & 2	                & Number of graph blocks \\
         Activation      	         & ReLu	            & ReLu              & ReLu	            & GELU	            & GELU	            & GELU	            & GELU	            & Activation function for MLP modules \\
         LR  	                     & 0.001	        & 0.001	            & 0.001	            & 0.001	            & 0.001	            & 0.001	            & 0.001	            & Learning rate \\ 
         LR Decay 	                 & 0.8	            & 0.8	            & 0.8	            & 0.8	            & 0.8	            & 0.8	            & 0.8	            & Learning rate decay after each epoch \\
         BS 	                     & 32	            & 32	            & 32	            & 32	            & 32	            & 32	            & 32	            & Batch size \\
         Dropout                     & 0.05	            & (0.05, 0.05, 0.15)$^*$ & 0.05	        & 0.05	            & 0.05	            & 0.05	            & 0.05	            & Dropout rate \\
         LSTM/MLP layers 	         & 2	            & (1, 1, 2)$^*$     & -	                & -	                & -	                & -	                & -	                & LSTM/MLP layers in each update function \\
         Heads 	                     & -	            & -	                & 8	                & 8	                & 8	                & 8	                & 8	                & Number of attention heads \\
         Attn Kernel 	             & -	            & -	                & -	                & 6	                & -	                & -	                & 3	                & Kernel for convolutional attention, see: \cite{li2019enhancing}\\
         Local Attn	                 & -	            & -	                & -	                & 4	                & -	                & -	                & -	                & Local attention length, see: \cite{li2019enhancing} \\
         Restart Length              & -	            & -	                & -	                & 16                & -	                & -	                & -	                & Restart attention length, see: \cite{li2019enhancing} \\
         Sparse Attn                 & -                & -                 & No                & Yes               & Yes               & Yes               & Yes               & Some sparse attention pattern \cite{li2019enhancing, wu2021autoformer, zhou2021informer} \\
         c 	                         & -                & -                 & -                 & -                 & 3                 & 3                 & 3                 & Sampling factor, see:  \cite{zhou2021informer, wu2021autoformer} \\
         mvavg kernel                & -	            & -	                & -	                & -	                & -	                & 25	            & -	                & Moving average for series decomposition \cite{wu2021autoformer} \\
         num decomp                  & -                & -                 & -                 & -                 & -                 & -                 & 4                 & Number of layers in MDWD, see Sec.~\ref{sec:theo:wt} \\ 
         Time (s)              & 0.0021	        & 0.0051	        & 0.0047	        & 0.0054	        & 0.0063	        & 0.0096	        & 0.0172	        & Average compute time for a 6-step forecast \\
         \hline
         \multicolumn{9}{l}{$^*$ Parameter values used for the 1, 6 and 24 step forecast models. Values not shown in parenthesis were kept the same for all horizons.  }
    \end{tabular}
    }
    \label{tab:parameters}
\end{table*}

All features were scaled separately using a standard scaler, with zero mean and unit variance.
Three forecasting horizons were considered for the multi-step forecasting problem, namely 1-, 6- and 24-step forecasts, corresponding to 10-minute, 1-hour and 4-hour periods.
Since the study considered multi-step forecasting, a prediction was made for every 10-minute interval also in the 1- and 4-hour ahead settings, instead of simply average wind speed forecasts over the respective periods. 
Models were trained to minimise the mean squared error (MSE) and hyperparameter tuning was conducted in Optuna \cite{akiba2019optuna}. 
All experiments were conducted in PyTorch on a single Nvidia RTX 2070 GPU.
Every model was trained for 30 epochs using an Adam optimiser.
Both look-back windows and model-specific parameters were treated as hyperparameters and decided based on the tuning procedure and computational constraints, with final model-parameters given in Table~\ref{tab:parameters}.
From the tuning, it was found that a look-back window of 32 time-steps was suitable for the 10-minute and 1-hour ahead forecasts, increased to 64 for the 4-hour forecasts, which seemed reasonable, as longer contexts might be useful to better predict trends further into the future.
Some explanation of the different parameters are given in Table 1, though the authors refer the interested reader to the provided references in the `Explanation' column for more detailed descriptions of model-specific parameters. 
In Table~\ref{tab:parameters}, $d$ are the model dimensionalities, while $d_\text{hidden}$ and  `Activation', refer to the dimensionality and activation function for hidden layers in the ST-MLP model and MLP modules in Transformer-based models.
All Transformer-based models employed the same architecture for the edge-update functions, $\phi^{(e)}$, namely a single vanilla Transformer encoder layer \cite{vaswani2017attention}.

For the ST-MLP and ST-LSTM models, both autoregressive and vector-output models were studied, where different training strategies such as mixed teacher forcing was studied for the autoregressive architectures. 
However, vector-based architectures were found to yield the best results for both the ST-MLP and ST-LSTM models, where the multiple forecasting steps are mapped directly using a linear projection with $P$ outputs corresponding to the forecast locations, as
\begin{align}
    h_{t} &= \text{LSTM}(X_{t}, X_{(t-1)}, ... X_{(t-(S-1))}) \\
    \hat{V} &= LT(h_{t}).
\end{align}
$X_t$ are the input features at time, $t$, respectively. $LT$ represents the linear projection that projects the last hidden features, $h_{t} \in \mathrm{R}^{N\times d}$, to the $P$ number of predicted outputs, $\hat{V} \in \mathrm{R}^{N\times P}$, as $LT: \mathbb{R}^d \to \mathbb{R}^P$, where $d$ is the hidden dimensionality of the model. 

The ST-LSTM model was quite sensitive to model parameters, explaining the different dimensionalities, $d$ in Table~\ref{tab:parameters}, used for this model. 
Instead of a simple linear projection, it was found beneficial to use a two-layer MLP as the final head for the ST-LSTM model, with parameters given by $d$, $d_{\text{hidden}}$ and `Activation' in Table~\ref{tab:parameters}.

Considering the ST-FFTransformer, the MHA modules in the trend (right-hand) stream in Fig.\ref{fig:fftrans}, employed convolutional \textit{ProbSparse} attention, but different to the original \textit{ProbSparse Attention}, which focuses on dominant queries, it was found marginally beneficial to instead locate dominant keys.
Convolutional \textit{ProbSparse Attention} was also used for the MHA module in the FFT-Attention, but finding top queries, as in the original formulation.
For non-dominant query locations, the original \textit{ProbSparse Attention} formulation assigns mean values to the outputs \cite{zhou2021informer}. 
However, when using the \textit{ProbSparse Attention} for the FFT-Attention, we instead set the outputs for non-dominant locations to zero to introduce sparsity and only select a subset of the possible frequencies. 
This was thought desirable, as to avoid outputs with a very large number of frequencies of small or similar amplitudes, which could be considered noisy. 

Considering the computational time for each model, it can be seen from the last row in Table~\ref{tab:parameters}, that the ST-FFTransformer model was slower than the other models. 
This was mainly due to the MDWD layer, which was not optimised to run in parallel over the different input features. 
Furthermore, both the ST-Autoformer and ST-FFTransformer models rely on computing FFTs, which are fairly slow and explain why these took longer to compute forecasts. 
However, since the time to compute any forecasts was much smaller than one second, and considerably less than the 10-minute sampling rate of the data, this was not considered a significant problem for this study and is therefore left for future work where computational speed might be more important. 

The persistence model was used as a benchmark against which to compare all the other models.
For the spatio-temporal setting, the persistence model will use the last recorded wind speed for a station as the forecast over the entire horizon, as $\hat{V}_{(t+1)} = \hat{V}_{(t+2)} = ... = \hat{V}_{(t+P)} = V_{t}$, using the notation in Sec.~\ref{sec:meth:data}.
Even though this is quite a trivial method for making forecasts, the model can achieve fairly accurate results in the short-term and is therefore used as an important baseline to outperform.

\section{Results and Discussion}\label{sec:res}

\subsection{Forecasting Error}\label{sec:res:error}
\begin{table*}[]
    \small
    \centering
    \captionsetup{justification=raggedright,singlelinecheck=false}
    \caption{Test losses in $m/s$ for different models and forecast horizons.}
    \resizebox{\textwidth}{!}{%
    \begin{tabular}{lcrcrcrcrcrcr}
        \hline
         Model              & \multicolumn{6}{c}{MSE (\% Improvement)}                                                         & \multicolumn{6}{c}{MAE (\% Improvement)}                                                            \\
         ~          & \multicolumn{2}{c}{1-step (10 min)}  & \multicolumn{2}{c}{6-steps (1 hrs)}  & \multicolumn{2}{c}{24-steps (4 hrs)} & \multicolumn{2}{c}{1-step (10 min)}  & \multicolumn{2}{c}{6-steps (1 hrs)}  & \multicolumn{2}{c}{24-steps (4 hrs)} \\
         \hline
        Persistence         & 0.4138    &  0.0 \%          & 1.0282    &   0.0 \%          & 2.9712    &   0.0 \%              & 0.4277    &   0.0 \%          & 0.6852    &  0.0 \%           & 1.1922    &  0.0 \%       \\
        ST-MLP              & 0.3938    &  4.8 \%          & 0.9108    &  11.4 \%          & 2.3951    &  19.4 \%              & 0.4265    &   0.3 \%          & 0.6588    &  3.9 \%           & 1.0987    &  7.8 \%       \\
        ST-LSTM             & 0.4040    &  2.3 \%          & 0.9430    &   8.3 \%          & 2.3707    &  20.2 \%              & 0.4336    &  -1.4 \%          & 0.6739    &  1.7 \%           & 1.0964    &  8.0 \%       \\
        ST-Transformer      & 0.4022    &  2.8 \%          & 0.8923    &  13.2 \%          & 2.3302    &  21.6 \%              & 0.4346    &  -1.6 \%          & 0.6561    &  4.3 \%           & 1.0866    &  8.9 \%       \\
        ST-LogSparse        & 0.3937    &  4.9 \%          & 0.8492    &  17.4 \%          & 2.3290    &  21.6 \%              & 0.4309    &  -0.7 \%          & 0.6369    &  7.0 \%           & 1.0838    &  9.1 \%       \\
        ST-Informer         & 0.3948    &  4.6 \%          & 0.8552    &  16.8 \%          & 2.2321    &  24.9 \%              & 0.4321    &  -1.0 \%          & 0.6383    &  6.8 \%           & 1.0559    &  11.4 \%       \\
        ST-Autoformer       & \textbf{0.3454} & \textbf{16.5 \%}    & 0.8203    & 20.2 \%          & 2.3990    & 19.3 \%              & \textbf{0.4046} & \textbf{5.4 \%}    & \textbf{0.6231} & \textbf{9.1 \%}    & 1.0907    & 8.5 \%       \\
        ST-FFTransformer    & 0.3492    & 15.6 \%          & \textbf{0.8147} & \textbf{20.8 \%}    & \textbf{2.1336} & \textbf{28.2 \%}        & 0.4098    & 4.2 \%          & 0.6245    & 8.9 \%           & \textbf{1.0329} & \textbf{13.4 \%} \\
        \hline
    \end{tabular}
    }
    \label{tab:errors}
\end{table*}

\begin{figure*}
    \begin{subfigure}{0.49\textwidth}
        \centering
        \includegraphics[width=0.75\textwidth]{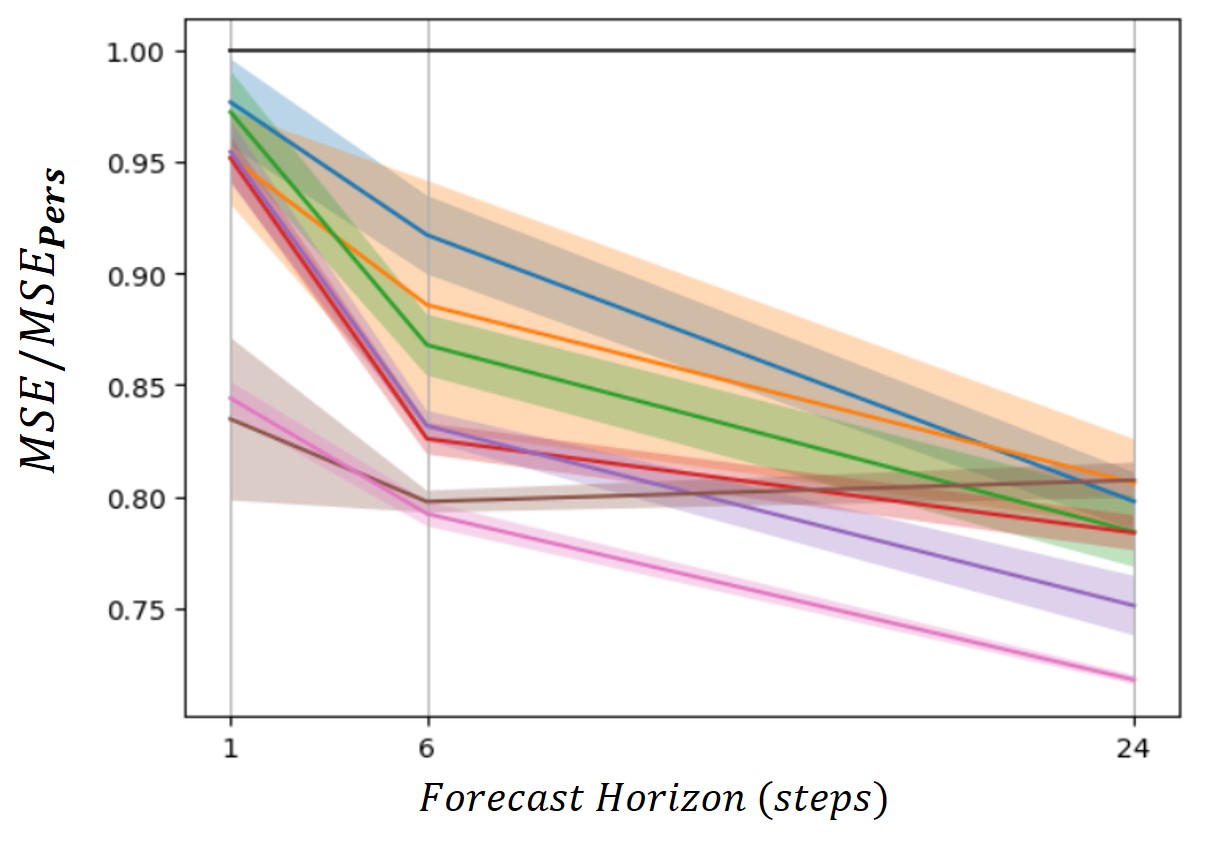}
        \caption{MSE divided by persistence MSE} \label{fig:res_mse}
    \end{subfigure}
    \hspace*{\fill}
    \begin{subfigure}{0.49\textwidth}
        \centering
        \includegraphics[width=0.75\textwidth]{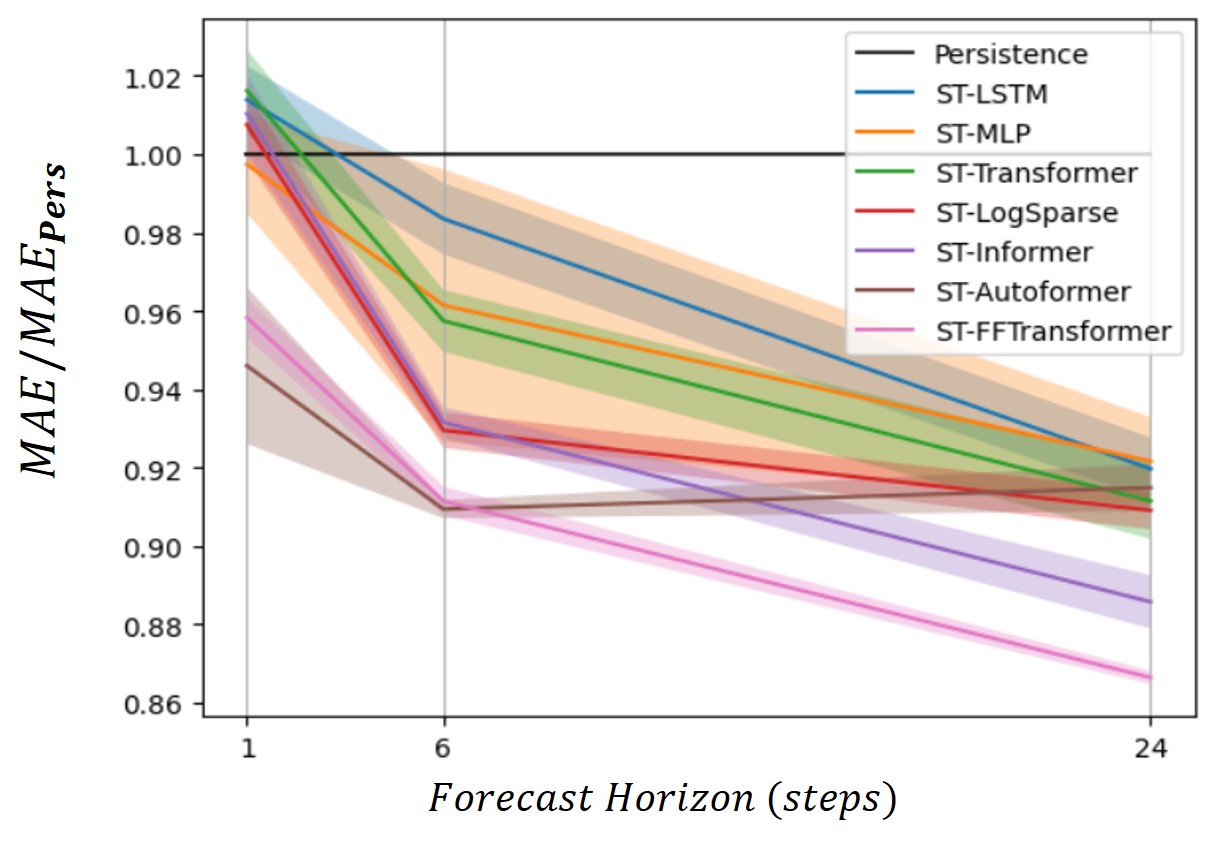}
        \caption{MAE divided by persistence MAE} \label{fig:res_mae}
    \end{subfigure}
\caption{Visualisation of the test errors for the different forecast horizons (1, 6 or 24 steps) divided by the respective persistence errors. Shaded areas show the $\pm$ one standard deviation from the five training iterations of each model.}
\label{fig:res_losses}
\end{figure*}

To evaluate the predictive performance of the different models, we start by comparing the mean absolute (MAE) and squared (MSE) errors, given by the following equations: 
\begin{align}
    \text{MAE} &= \frac{1}{n} \sum_{i=0}^n |y_i - \hat{y}_i| \\
    \text{MSE} &= \frac{1}{n} \sum_{i=0}^n (y_i - \hat{y}_i)^2 ,
\end{align}
where $n$ is the total number of samples and, $\hat{y}$, the model predictions, which should be close to the targets $y$.
For each forecasting horizon, every model was trained five times, with the average errors on the test set given in Table~\ref{tab:errors}.
Before computing the metrics, the predictions and labels were transformed back to a meters-per-second scale for better interpretability of the results. 
The percentage improvement values in Table~\ref{tab:errors} are relative to the persistence model and are provided in order to highlight the relative forecasting performances.
These are also visualised in Fig.~\ref{fig:res_losses}, where the shaded regions show the variability from the five training iterations of each model as $\pm$ one standard deviation.

Looking at the ST-Transformer model in Table~\ref{tab:errors}, it was evident that the model did not report any remarkable advancements.
Compared to the ST-MLP model, the ST-Transformer only showed marginal improvements for the 6- and 24-step forecasts, while being on-par with the persistence and ST-LSTM models for the single-step setting. 
For the 1- and 6-step forecasts, the ST-MLP model outperformed the ST-LSTM model, while the ST-LSTM only yielded slightly better results than the ST-MLP model for the longer 24-step forecasts. 
This might indicate that for the immediate short-term forecasts, all the previous time-steps might be relevant, resulting in the fully connected architecture of the ST-MLP achieving better accuracies than the ST-LSTM. 
However, for longer input sequences and forecast horizons, the LSTM architecture might be better at encoding the useful information, resulting in the improving performance of the ST-LSTM model, compared to the ST-MLP, for the longer forecasting settings. 
Furthermore, looking at Fig.~\ref{fig:res_losses}, the ST-MLP model showed significantly more variability between training iterations, compared to the other models. 
Since this variability was uncontrollable, it meant that one could not reliably conclude on the ST-MLP model's exact performance, which was undesirable.

The ST-LogSparse and ST-Informer performed consistently better than the ST-Transformer model across all forecasting horizons in terms of both MSE and MAE, which showed the potential improvements brought by the \textit{ProbSparse} and convolutional attention mechanisms for wind forecasting. 
Even though the ST-LogSparse and ST-Informer models reported slightly inferior forecasting performance in terms of MAE for the single-step forecasts, compared to the persistence model, both showed approximately a five percent improvement in MSE. 
Since MSE penalise large errors more heavily than the MAE metric, it meant that for the single-step forecasts, the persistence model had on average fewer slightly smaller errors, but a larger number of drastically wrong predictions than the ST-LogSparse and ST-Informer models.  

In general, all Transformer-based models attained better results than the ST-MLP and ST-LSTM models for the multi-step forecasts. 
The ST-Autoformer and ST-FFTransformer performed remarkably well for the 1- and 6-step forecasts, achieving more than three times the improvement reported for the third best model in terms of MSE for the 1-step forecasts, and nearly a doubling compared to the ST-MLP for the 6-step forecasts. 
Even though the ST-Autoformer model performed very well for the 1- and 6-step forecasts, its performance was seen to degrade for the 24-step setting, where it was inferior to all other Transformer-based models, clearly visualised in Fig.~\ref{fig:res_losses}. 
\begin{figure*}
    \begin{subfigure}{0.345\textwidth}
        \includegraphics[height=4.8cm]{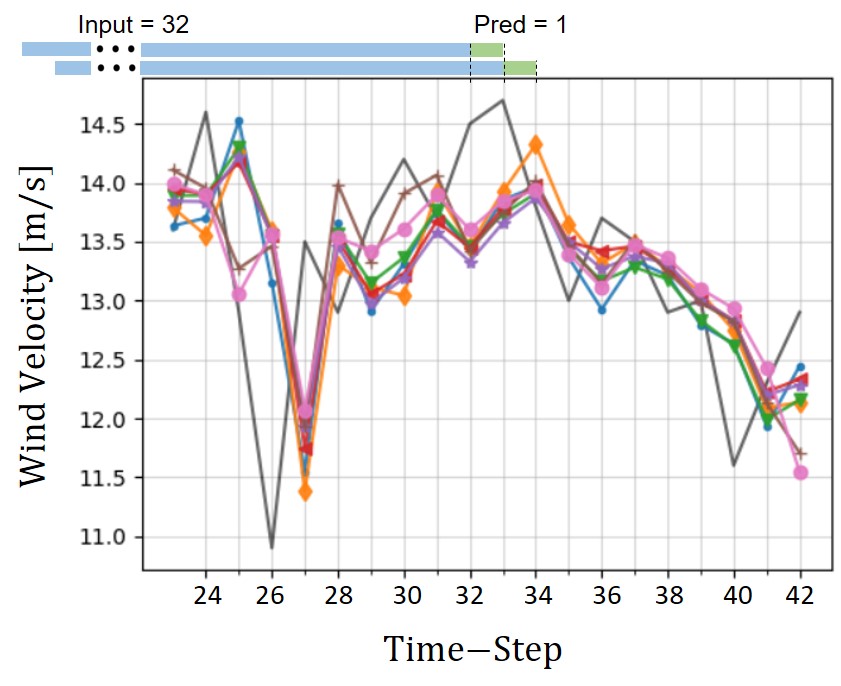}
        \caption{One-Step Forecasts (10-minute)} \label{fig:predictions_a}
    \end{subfigure}
    \hspace*{\fill}
    \begin{subfigure}{0.307\textwidth}
        \includegraphics[height=4.8cm]{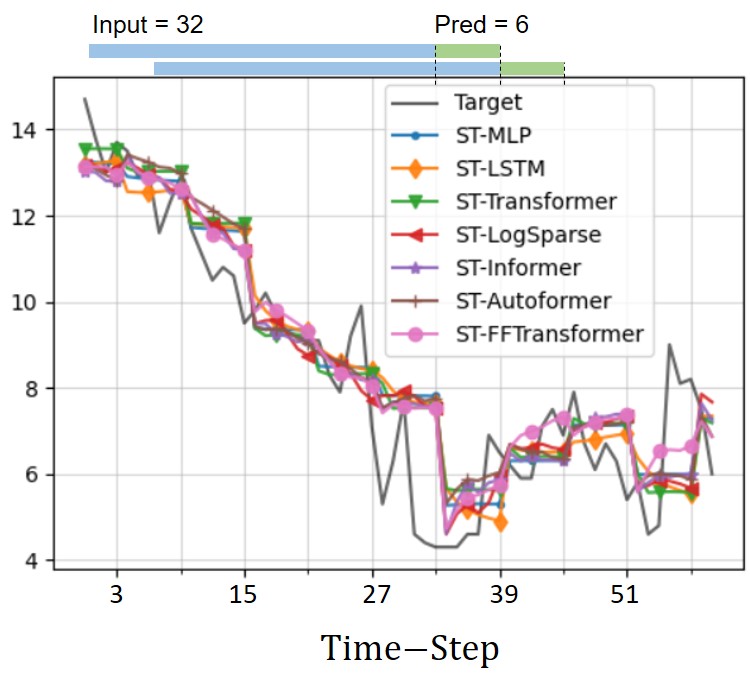}
        \caption{Six-Step Forecasts (1-hour)} \label{fig:predictions_b}
    \end{subfigure}
    \hspace*{\fill}
    \begin{subfigure}{0.307\textwidth}
        \includegraphics[height=4.8cm]{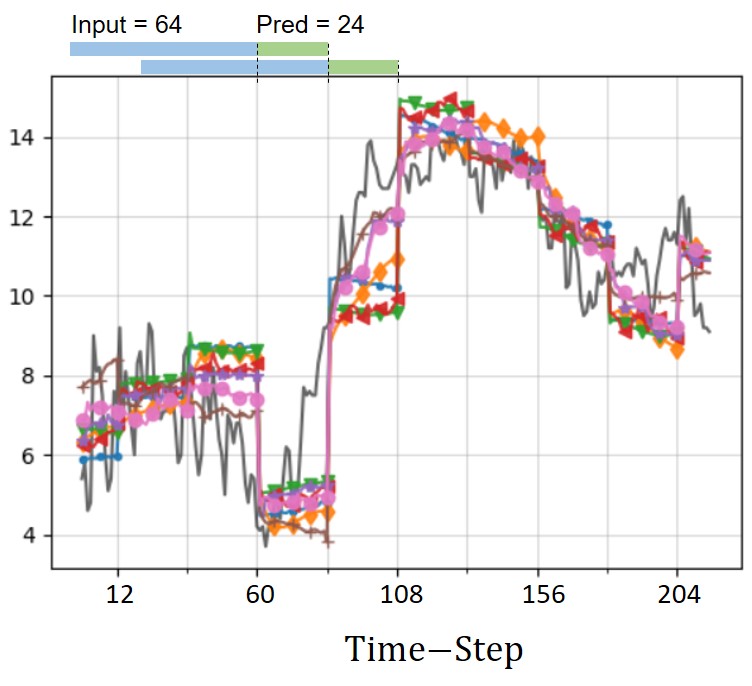}
        \caption{24-Step Forecasts (4-hour)} \label{fig:predictions_c}
    \end{subfigure}
\caption{Prediction examples of the different models under the 1-, 6- and 24-step forecasts. The vertical grid line spacings correspond to the prediction horizons, i.e. in (b), 6-step predictions are shown with vertical grid spacing 6.}
\label{fig:predictions}
\end{figure*}

The ST-FFTransformer on the other hand, continued to improve, achieving superior results compared to all other models also for the 4-hour forecasts. 
Both the ST-FFTransformer and ST-Autoformer achieved the best accuracy for three settings each, likely because of data decomposition. 
For the 1- and 6-step ahead forecasts, the difference between the two models was very small, while the ST-FFTransformer substantially outperformed the ST-Autoformer for the longer forecasting horizon of four hours.
In the short term, wind time series fluctuates rapidly, while for longer time periods, it might become more important to learn trend components as well as oscillations. 
One reason for the superior performance of the FFTransformer architecture for the longer forecasting setting might be that it separately processes the frequency and trend components, using the two streams described in Sec.~\ref{sec:meth:fft}, with attention and feed forward modules applied to both. 
On the other hand, the Autoformer model is mainly focused on processing periodic components using the Auto-Correlation module and does not perform significant processing on extracted trend components after series decomposition, making it potentially struggle to learn longer-term trends that lack periodicity. 
Such a hypothesis was strengthened by the fact that all other models seemed to perform comparatively better for the longer forecasting setting, compared to the ST-Autoformer.
Because of the two streams used in the ST-FFTransformer, it is able to perform well for both short-term forecasts, where periodic behaviour is thought more important, as well as for the four-hour forecasts, where non-oscillating trend components become increasingly important. 
Even though the ST-FFTransformer achieved good results across different horizons, it did not outperform the ST-Autoformer in the short-term. 
This was despite more advanced decomposition using MDWD, which was initially thought better at extracting trend and periodic components at different frequencies, compared to the simple moving average operation used in the Autoformer.
In terms of decomposition, this could indicate that the MDWD is not necessarily superior to the `Series Decomposition', or more likely, that the repeated use of decomposition in the Autoformer might work slightly better than the single decomposition performed on the inputs in the ST-FFTransformer. 
For future research it would be interesting to see how the FFTransformer architecture would perform with Auto-Correlation modules in place of the FFT Attention blocks in Fig.~\ref{fig:fftrans} and experimenting with some data decomposition applied to every encoder or decoder layer.

Considering the 1-step forecasts, the ST-Autoformer model exhibited considerable variability between training iterations, compared to the ST-FFTransformer, seen by the shaded regions in Fig.~\ref{fig:res_losses}, which was undesirable.
We therefore argue for the competitive forecasting performance of our proposed FFTransformer architecture, which performed consistently well across all forecasting horizons and with limited variability. 

Fig.~\ref{fig:predictions} shows some predictions under the different forecast horizons, for some randomly chosen time periods for the Kvitebjørnfeltet measurement station in Fig.~\ref{fig:stations}.
Overall, it was difficult to conclude on major differences between the models based on the plotted time series.
Nevertheless, it was observed that for the 6-step setting, the ST-MLP and ST-Transformer models often gave near-constant predictions for all six time-steps, while the ST-LSTM, ST-LogSparse, ST-Informer, ST-FFTransformer and ST-Autoformer were able to produce slightly more diverse time series.
For the 24-step (4 hrs) setting, all models showed more near-constant predictions, as seen in Fig.~\ref{fig:predictions_c}, where some forecasts exhibit step-like changes at every prediction interval (i.e. 24 time-steps).
Forecasting accurate time series for longer horizons is challenging and the near-constant predictions indicate that the models sometimes struggled to confidently predict sharp increases or decreases in wind speeds. 
Having additional features and measurement stations or much deeper networks together with larger datasets, could potentially enable the models to more accurately learn these long-term changes. 
However, some of the near-constant predictions might also reflect the inherent stochasticity associated with wind time series, meaning that in some scenarios it might be inconceivable to attain accurate multi-step forecasts for longer horizons, based solely on previous time series and no physical information. 
Nevertheless, for other time-steps, the ST-LSTM, ST-FFTransformer, ST-Autoformer and ST-Informer models seemed much better than the ST-MLP, ST-Transformer and ST-LogSparse at capturing the overall trends within the intervals in Fig.~\ref{fig:predictions_c}.

\begin{table*}[t]
    \small
    \centering
    \captionsetup{justification=raggedright,singlelinecheck=false}
    \caption{Estimated power errors for different models and forecast horizons.}
    \resizebox{\textwidth}{!}{%
    \begin{tabular}{lcrcrcrcrcrcr}
        \hline
         Model              & \multicolumn{6}{c}{MAE [kW] (\% Improvement)}                                                         & \multicolumn{6}{c}{Interval MAE [kWh] (\% Improvement)}                                                            \\
         ~          & \multicolumn{2}{c}{1-step (10 min)}  & \multicolumn{2}{c}{6-steps (1 hrs)}  & \multicolumn{2}{c}{24-steps (4 hrs)} & \multicolumn{2}{c}{1-step (10 min)}  & \multicolumn{2}{c}{6-steps (1 hrs)}  & \multicolumn{2}{c}{24-steps (4 hrs)} \\
         \hline
        Persistence         & 166.1    &  0.0 \%          & 265.4    &   0.0 \%          & 459.2    &   0.0 \%              & 27.684    &   0.0 \%          & 228.8    &   0.0 \%           & 1604.2    &   0.0 \%       \\
        ST-MLP              & 165.1    &  0.6 \%          & 253.2    &   4.6 \%          & 422.0    &   8.1 \%              & 27.509    &   0.6 \%          & 209.4    &   8.5 \%           & 1387.0    &  13.5 \%       \\
        ST-LSTM             & 166.0    &  0.1 \%          & 258.5    &   2.6 \%          & 418.9    &   8.8 \%              & 27.669    &   0.1 \%          & 212.0    &   7.3 \%           & 1386.0    &  13.6 \%       \\
        ST-Transformer      & 166.4    & -0.2 \%          & 250.7    &   5.5 \%          & 417.7    &   9.1 \%              & 27.729    &  -0.2 \%          & 206.9    &   9.5 \%           & 1373.0    &  14.4 \%       \\
        ST-LogSparse        & 164.7    &  0.8 \%          & 245.5    &   7.5 \%          & 417.4    &   9.1 \%              & 27.453    &   0.8 \%          & 203.5    &  11.0 \%           & 1371.1    &  14.5 \%       \\
        ST-Informer         & 165.1    &  0.6 \%          & 245.7    &   7.4 \%          & 405.7    &  11.7 \%              & 27.519    &   0.6 \%          & 204.4    &  10.7 \%           & 1347.3    &  16.0 \%       \\
        ST-Autoformer       & \textbf{156.3} & \textbf{5.9 \%}    & \textbf{239.8}    & \textbf{9.7 \%}          & 416.9    & 9.2 \%              & \textbf{26.046} & \textbf{5.9 \%}    & \textbf{196.3} & \textbf{14.2 \%}    & 1383.0    & 13.8 \%       \\
        ST-FFTransformer    & 158.5   & 4.6 \%         & 239.9 & 9.6 \%    & \textbf{396.5} & \textbf{13.7 \%}        & 26.409    & 4.6 \%          & 197.4    & 13.7 \%           & \textbf{1300.6} & \textbf{18.9 \%} \\
        \hline
    \end{tabular}
    }
    \label{tab:errors_power}
\end{table*}

To investigate the physical interpretation of the forecasting results in relation to wind energy production, two additional MAE metrics were computed and provided in Table~\ref{tab:errors_power}, which correspond to the estimated errors in kW and kWh. 
Powers were estimated based on the NREL 5 MW reference wind turbine for offshore system development \cite{jonkman2009definition}.
By transforming the true and predicted wind speeds to powers using the reference turbine's power curve and then calculating the MAEs, the results show crude estimates for the average power errors using the different models. 
For the first metric in Table~\ref{tab:errors_power}, results were fairly similar to those discussed in Table~\ref{tab:errors}, but arguably more interpretable, in terms of understanding the consequence of differing predictive performances and potential risks associated with the proposed models. 

Instead of looking at the point-wise power difference between the true and predicted values, an operator might be primarily concerned with the total overall energy for a future time interval. 
For the Interval MAE, predictions were again transformed to power values, before the values associated with a particular forecast interval were summed.
Finally, since each step was associated with a 10-minute interval, summed values were divided by six, to convert them to the total energy estimates in kWh. 
Therefore, the Interval MAE provides an estimate of the difference between the predicted and true total energies produced for the relevant forecast horizon. 
The performance of different models was similar to previous results, but with all models showing greater improvements over the persistence model.
Considering the ST-FFTransformer, it managed to reduce the error by 300 kWh for the 4-hour forecasts, corresponding to a 19\% improvement over the persistence model. 
This was fairly notable, as it also meant an additional $\approx 5 \%$ improvement compared to most other models, and illustrated the potential cost savings and benefits of more accurate forecasting models. 

\begin{figure}
    \centering
    \includegraphics[width=0.48\textwidth]{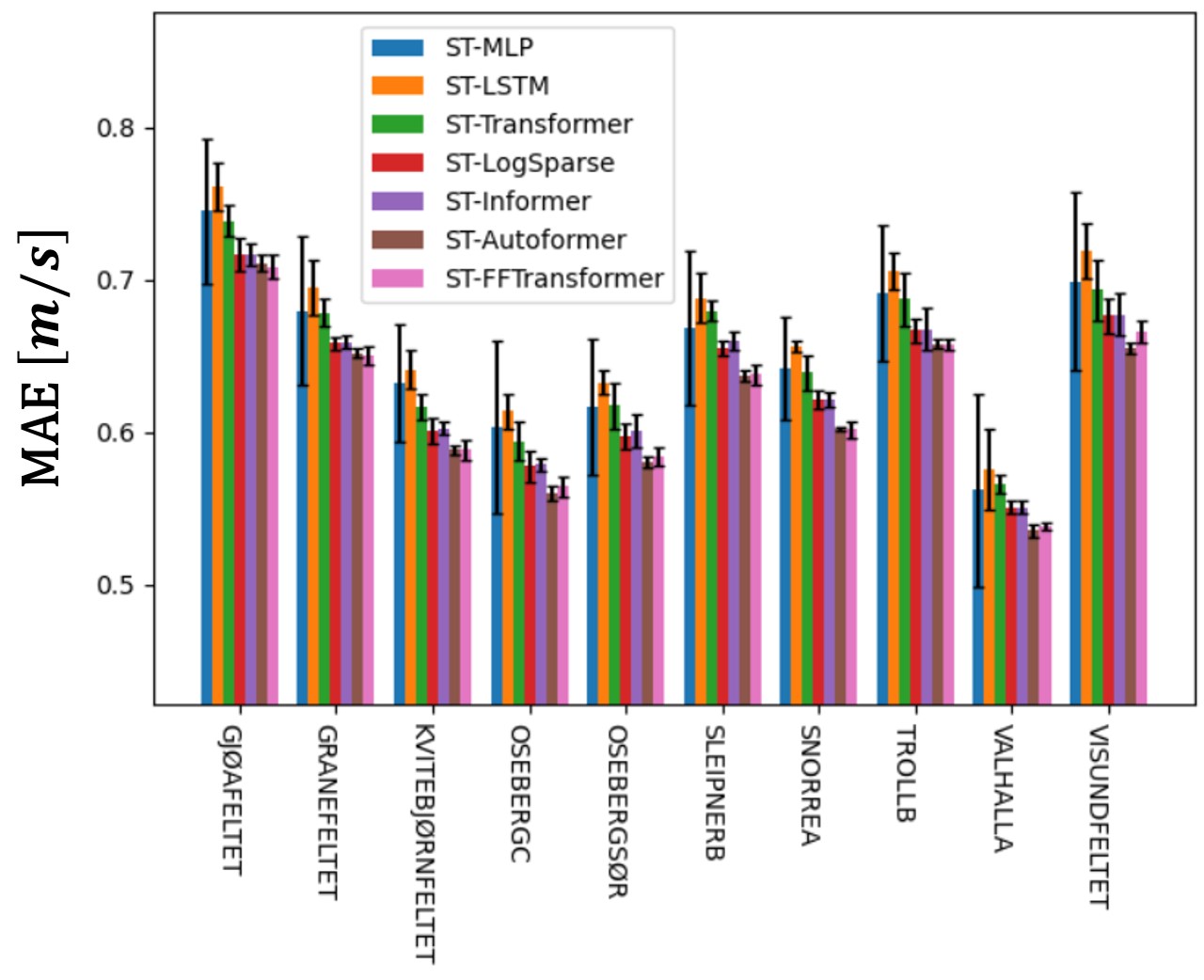}
    \caption{MAEs for the different available measurement stations in the test set under the 6-step forecasting setting. The error bars are the $\pm$ 2 standard deviations from the five training iterations of each model.}
    \label{fig:station_loss}
\end{figure}

\subsection{Station Predictability}\label{sec:res:station}

Since the models made forecasts for all 14 stations in Fig.~\ref{fig:stations}, simultaneously, it was thought interesting to inspect the average errors for each station independently.
Fig.~\ref{fig:station_loss}, shows the average MAEs for every station under the 6-step forecasting setting, along with error bars of $\pm 2\sigma$, where $\sigma$ is the standard deviation computed based on five separately trained models. 
It was seen that there was significant variability in how well the models were able to make forecasts for the different stations, with MAEs ranging from 0.53 $m/s$ for some stations to 0.74 $m/s$ for others. 
Even though there were distinct differences in the performance of different models, as discussed in the previous section, stations associated with higher or lower MAEs seemed consistent across all models.
The data only contained historic information on the specific measurement stations in Fig.~\ref{fig:stations}, with a fixed physical layout, meaning that spatial features would not change and that the meteorological data was likely to follow a particular distribution, specific to the area considered.
As a result, a station might be inherently easier to forecast than others, due to its location relative to surrounding stations and due to the dominant wind fields for this specific geographical area potentially being preferable for forecasting at a particular location.

\subsection{Graph Connectivity}\label{sec:res:connect}
\begin{figure*}
    \centering
    \includegraphics[width=0.9\textwidth]{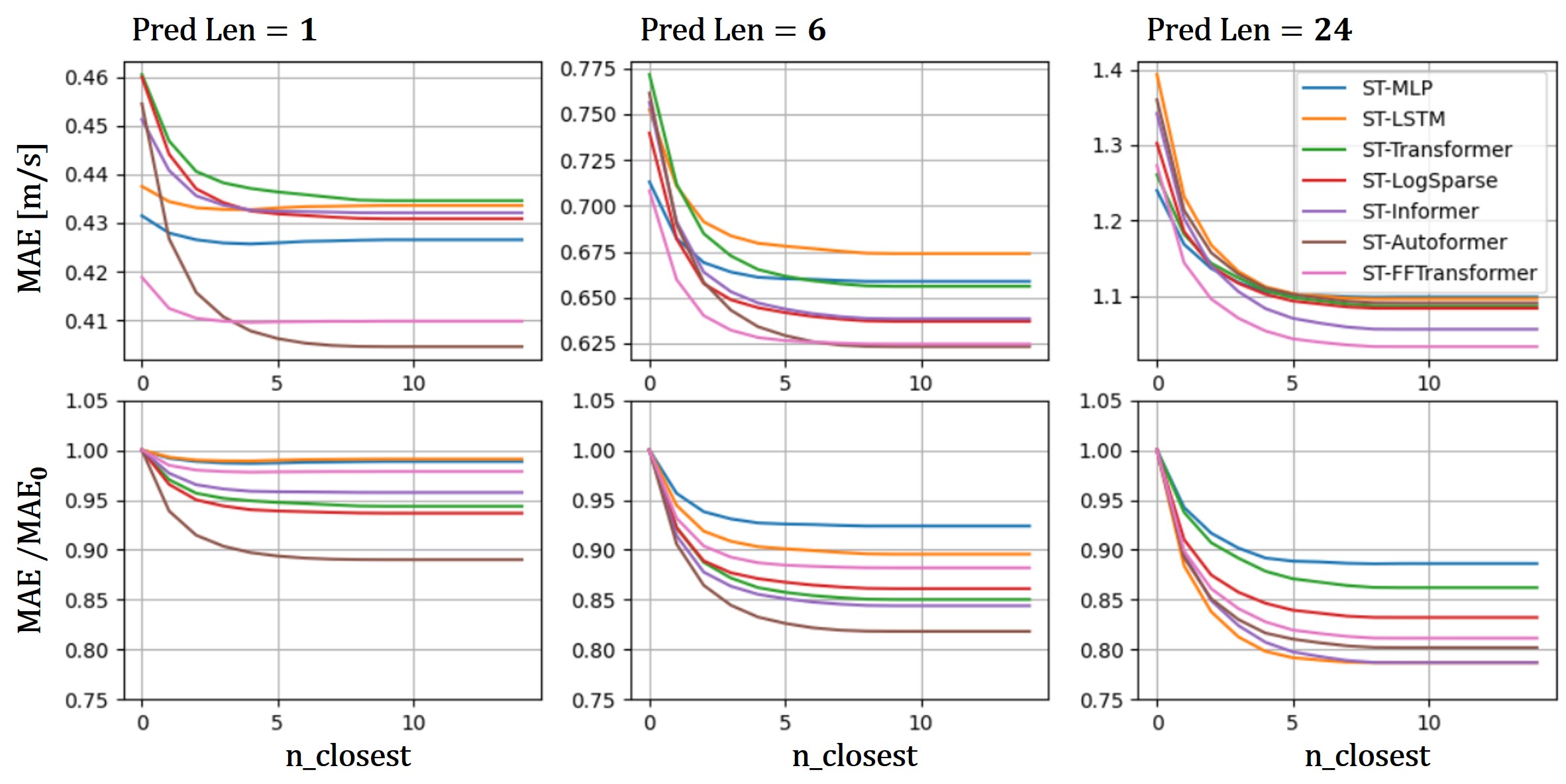}
    \caption{Test MAEs for different degrees of graph connectivity, with `n\_closest' referring to the maximum number of edges sending to a particular node. In the bottom row are the MAEs normalised by the MAEs for `n\_closest' = 0.}
    \label{fig:connect}
\end{figure*}

For the results discussed so far, all input data was constructed as complete graphs, meaning that all nodes had edge features sending to all other nodes in the network and itself. 
This trivial method for formulating the graphs could result in very large inputs, significantly increasing the computational and memory costs.
For instance, if a graph had 14 nodes, it would result in $14^2 = 196$ edges. 
Some of the edges might be redundant, meaning that the relevant spatial information could be provided by a subset of the edges, in addition to excess information potentially making training more challenging. 
Even though this study did not focus on discussing better connectivity strategies for wind forecasting or using learnable adjacency matrices, we conduct a brief investigation into whether some of the connections could potentially be removed. 
Fig.~\ref{fig:connect} plots the different MAEs on the test data, as we increase the number of connections in the input graphs. 
`n\_closest' refers to the number of closest neighbours to which we allow a node to connect. 
Looking at Fig.~\ref{fig:stations}, this meant that if `n\_closest' was set to three, Valhall A would only have edges sending to it from Sleipner B, Granefeltet and Oseberg Sør. 
The first row in Fig.~\ref{fig:connect}, shows the MAEs for different `n\_closest' values, while the second row contains the same information, but normalised by MAE$_0$, which was the MAE when `n\_closest' $= 0$, i.e. the MAE for predictions made without any spatial information.
While the first row provides information on the relative differences between models, the second row visualises the percentage improvements gained from including additional edges for a particular model. 

First, MAEs were seen to rapidly decrease as we increased the number of edges, before converging to constant values when more than around five neighbours were considered for the edges. 
The sharp decrease indicated that the models were able to successfully leverage spatial correlations to improve forecasts, proving the effectiveness of the proposed GNN architectures. 
Nevertheless, since MAEs converged to constant values for non-complete graphs, it indicated that a number of connections could potentially be removed without impairing predictive performance.
Further work would therefore be desirable to investigate better methods for which to construct the graphs or learn optimal connectivity for spatio-temporal wind forecasting.

Looking at the second row of Fig.~\ref{fig:connect}, it was seen that the percentage reduction in MAEs was greater for the longer forecasting horizons. 
For the immediate short-term (i.e. prediction length of 1), spatial correlations were thought less important than for the 1- and 4-hour forecasts, due to the large physical distances between nodes resulting in wind fields not having time to propagate in the immediate short-term. 
The added benefit of having long-range connections between nodes far apart, was therefore greater for the 6- and 24-step settings, than for the 1-step ahead forecasts. 
Comparing the 6- and 24-step forecasts, the latter also converged slightly later, which was likely due to the longer-term forecasts taking advantage of information from nodes further away from the target. 
The percentage change in MAEs was also greater for the 24-step setting than for the 6-step, even though the difference was not as big as between the 1- and 6-step settings, which indicated that the 24-step forecasts might have benefited from even larger geographical information than was available. 
Overall, it was concluded that all models were able to take advantage of spatial information in making forecasts, with the Transformer-based architectures generally showing slightly larger improvements than the ST-MLP model.

\section{Conclusion}
In recent years, Transformer-based models have presided over sequence-based deep learning, often superseding recurrent or convolutional models. 
Nevertheless, research employing these architectures for wind forecasting has been scarce. 
This study considered different Transformer architectures as the main predictor for spatio-temporal multi-step forecasting, focusing on the LogSparse Transformer, Informer and Autoformer. 
This is the first time many of these have been applied to wind forecasting and placed in a spatio-temporal setting using GNNs. 
Additionally, the novel FFTransformer was proposed, which is based on signal decomposition using wavelet transform and an adapted attention mechanism in the frequency domain. 
Results show that the FFTransformer architecture was very competitive, achieving results on par with the Autoformer-based model for the 1- and 6-step forecasts, while significantly outperforming all other models for the longer 24-step forecasts. 
Even though the vanilla Transformer architecture generally did not yield significant improvements over an MLP model, it was seen that the convolutional attention in the LogSparse Transformer and the \textit{ProbSparse Attention} of the Informer, were able to slightly improve prediction performance.
By estimating the associated prediction errors in kW and kWh, we showed the potential physical effects of different forecasting performances with regards to the power grid, with the FFTransformer model showing an additional 5 \% improvement over all other models for the 4-hour forecasts.
Nevertheless, obtaining the powers based on the NREL 5 MW reference turbine, the method was fairly trivial and it would be desirable to further test the different models on real wind power datasets. 
By removing graph connections in the input data, we showed that the proposed GNN architectures were successful in leveraging spatial correlations to improve local forecasts. 
However, it was also seen that some connections in the input data might be redundant, calling for additional research into more efficient approaches for graph connectivity in the context of wind forecasting. 
Since the FFTransformer model is not restricted to a particular signal decomposition technique or attention mechanism, slight alterations from the particular set-up used in this study might also be relevant and could be easily implemented and tested to facilitate different applications. 
We therefore hope that this study will spark further research into modifications and other applications of the FFTransformer, as well as investigation into the applicability of different Transformer-based architectures for use in wind forecasting.

\section*{CRediT authorship contribution statement}
\textbf{Lars Ø. Bentsen:} Project administration, Conceptualization, Data curation, Formal analysis, Investigation, Methodology, Project administration, Resources, Software, Validation, Visualization, Writing - original draft.
\textbf{Narada Dilp Warakagoda}: Supervision, Writing - review/editing.
\textbf{Roy Stenbro}: Supervision. Writing - review/editing.
\textbf{Paal Engelstad}: Supervision. Writing - review/editing.

\section*{Acknowledgements}
This work was in part financed by the research project ELOGOW (Electrification of Oil and Gas Installation by offshore Wind). 
ELOGOW (project nr: 308838) is funded by the Research Council of Norway under the PETROMAKS2 program with financial support of Equinor Energy AS, ConocoPhillips Skandinavia AS and Aibel AS.
The authors would also like to thank The Norwegian Meteorological institute for enabling the open access data used for this study. 

%Bibliography
\bibliographystyle{unsrt}  
\bibliography{references}

\end{document}